%% file: main.tex
\documentclass[10pt,twocolumn,letterpaper]{article}

\usepackage[pagenumbers]{wacv} 


\definecolor{wacvblue}{rgb}{0.21,0.49,0.74}
\usepackage[pagebackref,breaklinks,colorlinks,allcolors=wacvblue]{hyperref}
\usepackage{comment}
\usepackage{multirow, tabularx}
\usepackage{pifont}
\newcommand{\cmark}{\ding{51}}%
\newcommand{\xmark}{\ding{55}}%
\def\wacvPaperID{599} 
\def\confName{WACV}
\def\confYear{2026}

\title{\textbf{\textit{StyleYourSmile}}: Cross-Domain Face Retargeting Without Paired Multi-Style Data}

\author{Avirup Dey\\
University of Bath\\
Bath, United Kingdom\\
{\tt\small ad2837@bath.ac.uk}
\and
Vinay Namboodiri\\
University of Bath\\
Bath, United Kingdom\\
{\tt\small vpn22@bath.ac.uk}
}

\begin{document}
\twocolumn[{%
\renewcommand\twocolumn[1][]{#1}%
\maketitle
\vspace{-2.0cm}
\begin{center}
    \centering
    \captionsetup{type=figure}
    \includegraphics[width=0.9\textwidth]{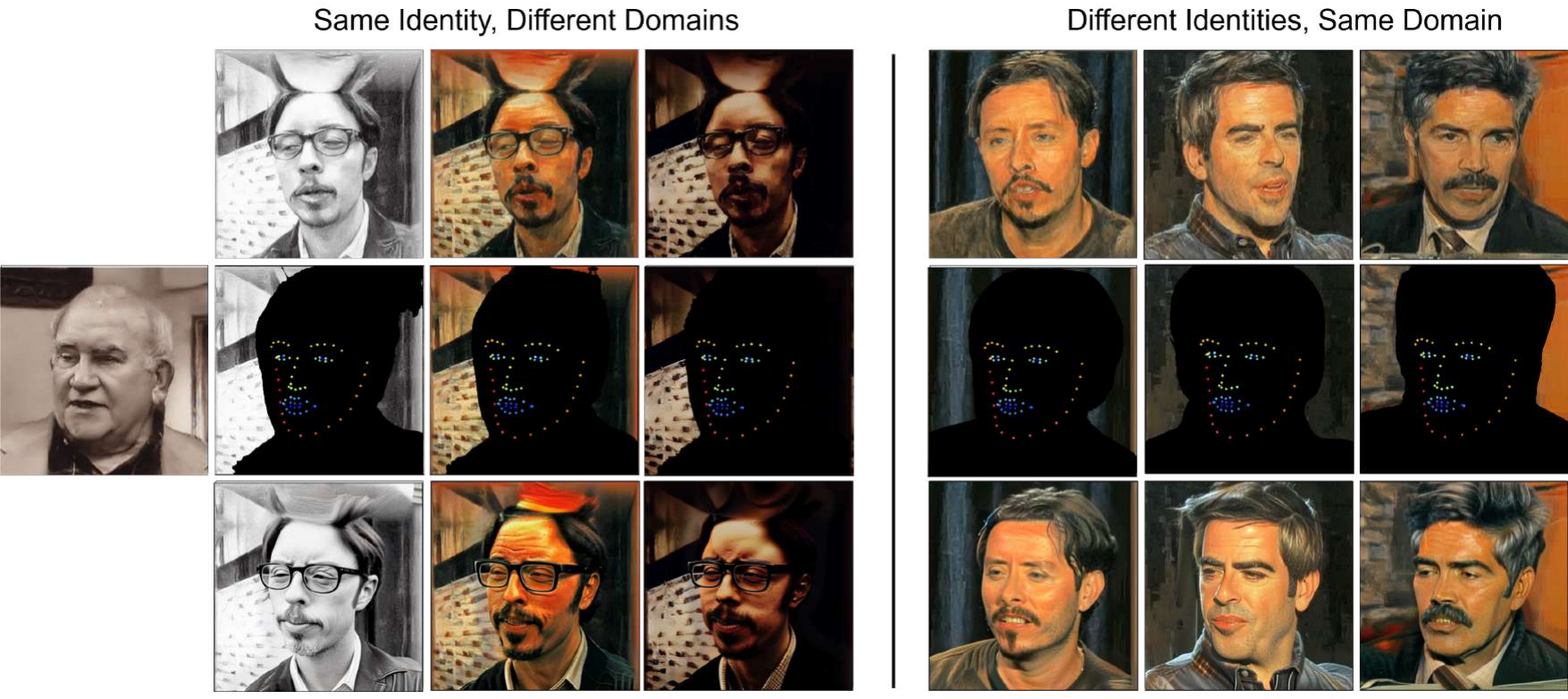}
    \caption{\textit{\textbf{StyleYourSmile}} can preserve fine-grained identity features as well as domain-specific attributes while retargeting facial expressions. Our model achieves disentanglement between identity and domain style \textbf{\textit{without}} using any curated multi-style pairs.}
    \label{fig:title}
\end{center}%
}]
\maketitle
\input{sec/0_abstract.tex}    
\input{sec/1_intro.tex}
\input{sec/2_related.tex}
\input{sec/3_method.tex}
\input{sec/4_experiments.tex}
\input{sec/5_ablations.tex}
\input{sec/6_conclusion.tex}
{
    \small
    \bibliographystyle{ieeenat_fullname.bst}
    \bibliography{main.bib}
}
\input{sec/suppl.tex}

\end{document}

%% file: sec/0_abstract.tex
\begin{abstract}
Cross-domain face retargeting requires disentangled control over identity, expressions, and domain-specific stylistic attributes. Existing methods, typically trained on real-world faces, either fail to generalize across domains, need test-time optimizations, or require fine-tuning with carefully curated multi-style datasets to achieve domain-invariant identity representations. In this work, we introduce \textit{StyleYourSmile}, a novel one-shot cross-domain face retargeting method that eliminates the need for curated multi-style paired data. We propose an efficient data augmentation strategy alongside a dual-encoder framework, for extracting domain-invariant identity cues and capturing domain-specific stylistic variations. Leveraging these disentangled control signals, we condition a diffusion model to retarget facial expressions across domains. Extensive experiments demonstrate that \textit{StyleYourSmile} achieves superior identity preservation and retargeting fidelity across a wide range of visual domains.
\end{abstract}

%% file: sec/1_intro.tex
\section{Introduction}
\label{sec:intro}
Imagine taking an old photograph of your loved one and transforming it to match your artistic vision—making the portrait smile exactly as you once sketched in your notebook! As trivial as it seems, this is a particularly challenging problem, even for state-of-the-art generative models. \par
Entanglement between identity and transient attributes like expression, pose, lighting, is the core challenge in facial image synthesis. Previous works \cite{kim2018deep, bounareli2023hyperreenact, GIF2020} leverage parametric face models \cite{li2017learning, giebenhain2023learning} to decouple identity from transient attributes. However, there is little research on preserving this disentanglement across domains (like photo $\rightarrow$ painting). Adding style as another orthogonal axis of variation poses an unconventional difficulty for existing methods, because (i) such domain shifts introduce structural inconsistencies— a dimple in a photograph may manifest as an intensity gradient, while in a painting, it could be a few abstract strokes and (ii) lack of multi-style datasets makes training nearly intractable. In this paper, we propose an image-based cross-domain face retargeting approach that can provide one-shot inference abilities and can be trained without curated multi-style paired data. \par
\input{sec/tables/teaser_tab.tex}
Recent works based on video diffusion \cite{drobyshev2024emoportraits, wei2024aniportrait} demonstrate strong identity retention across disparate visual domains however they rely on a single `ReferenceNet' \cite{hu2024animate} to encode identity and appearance directly from the source image. Despite the simplicty of this approach, these models are often trained at scale with very large datasets spanning multiple domains, including various animated, CGI clips \cite{ma2024follow}.\par 
Methods that use compressed representations like facial recognition embeddings \cite{papantoniou2024arc2face, wang2024instantid} achieve strong identity retention with smalled compute budget but discard non-identity details (e.g., style, lighting, accessories) due to the highly discriminative nature of these embeddings, as illustrated in Fig \ref{fig:arc2face}. 
\input{sec/figures/arc2face.tex}
While such stripped-down representations is useful for identity retention alone, it becomes a bottleneck for face retargeting, which demands preserving more nuanced attributes. In cross-domain scenarios preservation of domain-specific cues is critical. Hence, we propose two complementary  encoders - a face recognition-based encoder that learns identity features agnostic to domain style, and a style encoder that complements it by learning representations for domain-specific cues. The spatial conditioning needed for accurate retargeting is generated by a 3DMM and routed to the diffusion model via ControlNet. \par
Previous methods like HyperReenact \cite{bounareli2023hyperreenact} and GIF \cite{GIF2020} trains StyleGAN with outputs of parametric face models \cite{li2017learning} such that the latent is `pulled' towards the desirable pose without identity drift. However, being trained on real-world images only these cannot preserve fine-grained details in other domains and often produces over-smoothed images as shown in Fig. \ref{fig:qual1}. DiffusionRig \cite{ding2023diffusionrig} too is limited by its single-frame real-world dataset \cite{karras2019style} Hence, we also propose a light-weight style augmentation module that multiplexes standard video datasets like VoxCeleb \cite{nagrani2020voxceleb} with visual styles of different domains, as shown in Fig. \ref{fig:augment}. \par
We jointly fine-tune the image encoders and the diffusion model on the augmented dataset. Since our primary goal is to adapt the diffusion model to abstract styles, low rank fine-tuning \cite{hu2022lora} of the UNet suffices. The architecture choice and training formulation helps us achieve \textbf{\textit{one-shot cross-domain face retargeting}}. In summary, our key contributions are:
\begin{itemize}
    \item We propose \textbf{\textit{StyleYourSmile}}, a unified framework for one-shot cross-domain face retargeting with image-based diffusion models.
    \item We propose a novel training pipeline that (a) sidesteps curated multi-style data by augmenting real-world images using a fast, training-free technique and (b) disentangles identity and domain-specific style with two encoders that capture complementary representations. Our model can be trained using significantly lower compute budget compared to video-diffusion models (4x NVIDIA A5000 GPUs).
    
    \item Through extensive comparisons on VoxCeleb1 \cite{nagrani2020voxceleb}, a large scale face dataset, we demonstrate our model's superior performance against existing baselines for cross-domain retargeting.
\end{itemize}

%% file: sec/tables/teaser_tab.tex
\begin{table}[]
\centering
\resizebox{\linewidth}{!}{%
\begin{tabular}{l|ccc}
\hline
                      & \multicolumn{1}{l}{\textbf{Cross-Domain}} & \multicolumn{1}{l}{\textbf{Retargeting}} & \multicolumn{1}{l}{\textbf{One-shot}} \\ \hline
\textbf{HyperReenact} \cite{bounareli2023hyperreenact}   & \xmark                                         & \cmark                                        & \cmark                                     \\
\textbf{DiffusionRig} \cite{ding2023diffusionrig} & \xmark                                         & \cmark                                        & \xmark                                     \\
\textbf{Arc2Face} \cite{papantoniou2024arc2face}  & \xmark                                         & \xmark                                        & \cmark                                     \\
\textbf{Ours}         & \cmark                                         & \cmark                                        & \cmark                                     \\ \hline
\end{tabular}%
}
\caption{Scope of various face synthesis models: Unlike previous models, \textbf{\textit{StyleYourSmile}} is trained for one-shot high fidelity \textbf{\textit{cross-domain}} face retargeting.}
\label{tab:teaser_table}
\end{table}

%% file: sec/figures/arc2face.tex
\begin{figure}
    \centering
    \includegraphics[width=0.8\linewidth]{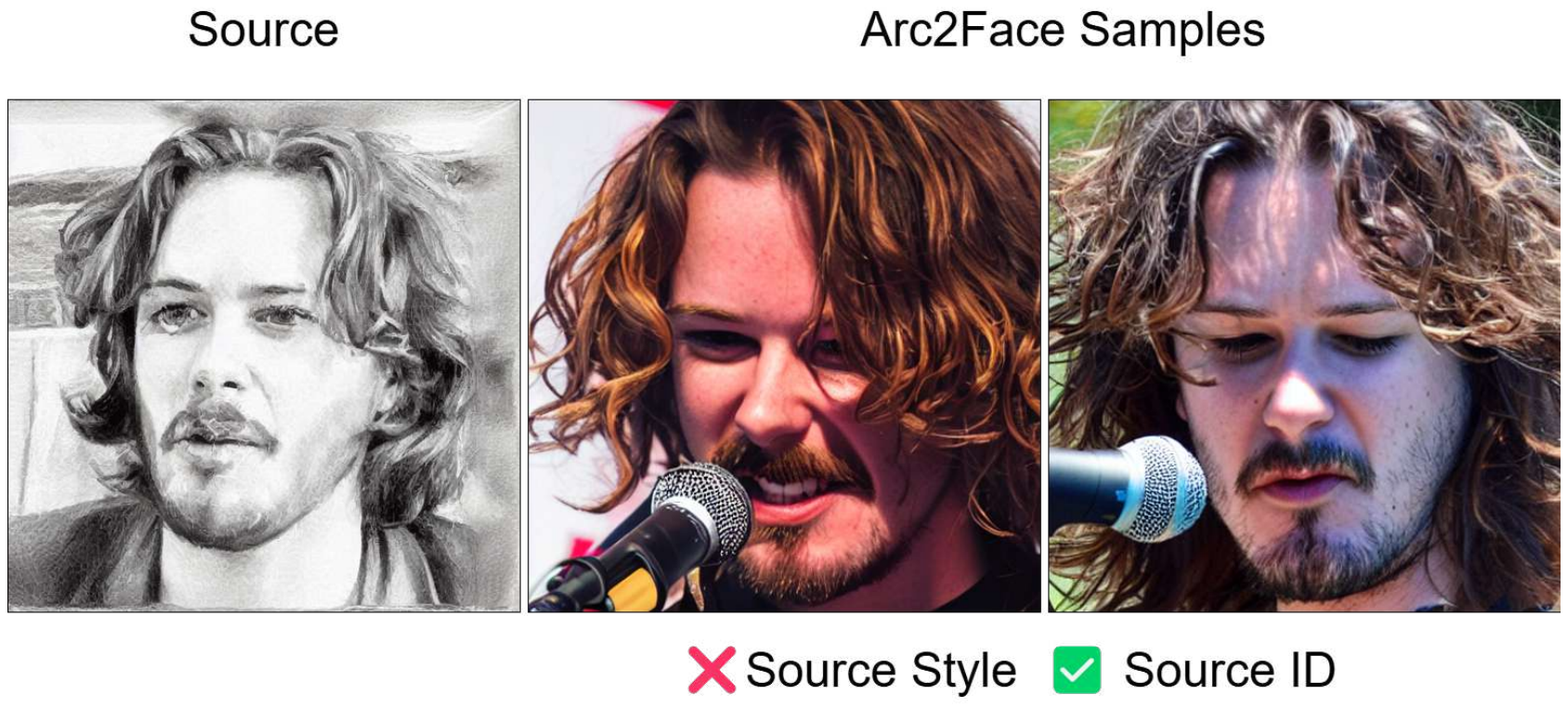}
    \caption{Arc2Face has strong identity retention but it cannot preserve the source style as the underlying face recognition encoder discards all information that is not relevant to a person's identity.}
    \label{fig:arc2face}
\end{figure}

%% file: sec/2_related.tex
\section{Related Works}
\label{sec:related}

\subsection{Identity-Aware Image Generation}
Identity-aware generation benefits significantly from face recognition models, which extract detailed identity features from facial images. These models are designed to capture comprehensive facial embeddings for identity similarity assessment \cite{wang2018cosface, deng2019arcface, boutros2022elasticface}. Consequently, inverting such models in a black-box manner has demonstrated the ability to generate facial images directly from identity embeddings \cite{mai2018reconstruction, razzhigaev2020black, vendrow2021realistic}, including applications in zero-shot generation using both GAN and diffusion architectures \cite{duong2020vec2face, kansy2023controllable}. However, existing inversion methods often rely on low-resolution datasets \cite{duong2020vec2face, truong2022vec2face} tailored for face recognition or employ high-quality but limited datasets, restricting their overall generalization capability.\par

Recent advancements have showcased impressive results through extensions of Stable Diffusion \cite{rombach2022high}. These include works based on Textual Inversion \cite{gal2022image} and DreamBooth \cite{ruiz2023dreambooth}, where diffusion models are fine-tuned on a few subject-specific images to learn a subject identifier, enabling accurate reproduction of that individual. Subsequent works have focused on reducing optimization time, such as HyperDreamBooth \cite{ruiz2024hyperdreambooth}, which integrates LoRA \cite{hu2022lora, yang2024lora} and a hypernetwork for fine-tuning from a single image. Other methods, like Celeb-Basis \cite{yuan2023celebbasis} and StableIdentity \cite{wang2024stableidentity}, learn an embedding basis from a celebrity dataset to condition text-based models. More broadly, Kosmos-G \cite{pan2023kosmos} introduced a multimodal perception model that processes diverse inputs, including facial images.\par

Related to face retargeting, there are advancements that condition diffusion models directly on facial features, enabling tuning-free personalization. Approaches such as FastComposer \cite{xiao2024fastcomposer}, PhotoVerse \cite{chen2023photoverse}, and PhotoMaker \cite{li2024photomaker} utilize CLIP image features to represent the input subject, although these methods are inherently limited by CLIP’s facial encoding capabilities. To enhance fidelity, models such as Face0 \cite{valevski2023face0}, DreamIdentity \cite{chen2024dreamidentity}, and PortraitBooth \cite{peng2024portraitbooth} incorporate face recognition embeddings as additional conditioning signals. IPAdapter \cite{ye2023ip} refines this process through a decoupled cross-attention mechanism that separates text and subject conditioning. InstantID \cite{wang2024instantid} further extends it by integrating an additional network for stronger identity guidance and facial landmark conditioning. Finally, FaceStudio \cite{yan2023facestudio} combines CLIP and ID embeddings, achieving remarkable stylized results. Arc2Face \cite{papantoniou2024arc2face} conditions a diffusion model solely on face recognition embeddings from a large-scale dataset \cite{zhu2021webface260m} for strong identity preservation. This, however, limits style preservation. \par
In contrast to the aforementioned works, our model is trained with two encoders that learn complementary representations for disentangling style from identity. This ensures preservation of source style and identity in cross-domain retargeting.

\subsection{Generative Models for Face Retargeting}
Retargeting models often utilize rendered images from 3DMM \cite{kirschstein2024diffusionavatars, yang2023diffusion, GIF2020} as motion control conditions. The facial prior from 3DMM makes the generation robust to large pose 
 and expression variations, but it provides only coarse facial texture and lack details for hair, teeth, and eye movement. StyleHEAT \cite{yin2022styleheat} and HyperReenact \cite{bounareli2023hyperreenact} leverage StyleGAN2 \cite{karras2020analyzing} to improve synthesis quality. However, StyleHEAT is constrained by its dataset of frontal portraits, while HyperReenact faces limitations in resolution and suffers from background blurring. Similarly, GIF \cite{GIF2020} focuses on disentangling facial attributes with FLAME \cite{li2017learning} for better controllability, but its reliance on latent space manipulation limits fine-grained detail synthesis.\par
Deep Video Portraits \cite{kim2018deep}, based on deferred neural rendering, pioneered high-quality video-driven facial retargeting by transferring head pose, expressions, and eye movements between subjects. However, it requires extensive data and computationally expensive training. ROME \cite{khakhulin2022realistic} adapt this framework for a single-image setting. While this enables avatar construction from a single frame, accurately capturing fine details from limited input remains a challenge.\par
Recently, diffusion models have gained prominence in face modelling. DiffusionRig \cite{ding2023diffusionrig} trains a DDPM \cite{ho2020denoising} model conditioned on FLAME \cite{li2017learning} buffers. Although it achieves disentangled control over facial attributes, it requires the user to fine-tune the model at inference, with a personal album. Video-diffusion based methods \cite{ma2024follow, wang2024v, wei2024aniportrait} offer better subject personalization across domains. The underlying ReferenceNet \cite{hu2024animate} captures a multi-resolution representation of the reference image which is fed to the attention layers of the denoising UNet. However, these methods are  difficult to operate without multiple high-end GPUs. Another work, FaceAdapter \cite{han2024face} enables one-shot face retargeting and face swapping. Unlike DiffusionRig it does not need test time optimisation. However, it has not been conditioned for cross-domain scenarios.
\par
In contrast to existing methods, our model is designed for one-shot face retargeting across domains without resorting to \textit{curated} multi-style data, heavy architectures and high-end GPUs. Our method is built on two key insights: (1) identity and style can be disentangled when a subject is observed across diverse styles and (2) pairing a discriminative identity representation with one that captures perceptual likeness encourages identity-style disentanglement.

\input{sec/figures/model.tex}

%% file: sec/figures/model.tex
\begin{figure*}
    \centering
    \includegraphics[width=0.7\linewidth]{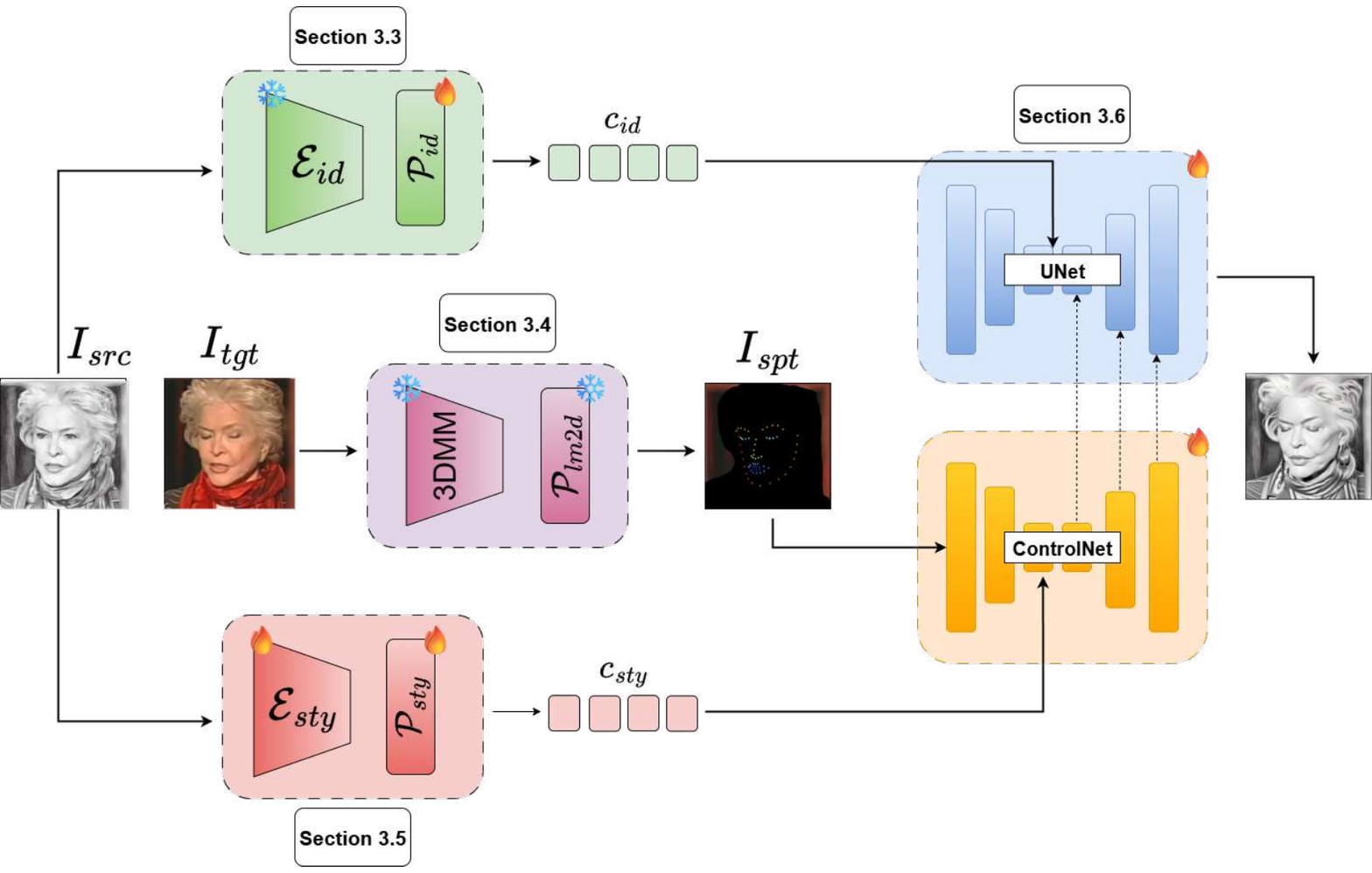}
    \caption{\textbf{Model Overview}: First, the source image $I_{src}$ is encoded as follows - (i) a face recognition encoder $\mathcal{E}_{id}$ extracts domain invariant indentity features and they are projected into CLIP text space by a decoder $\mathcal{P}_{id}$ as identity tokens $c_{id}$. (ii) A style encoder $\mathcal{E}_{sty}$ extracts domain specific style features and they are projected into CLIP text space by a decoder $\mathcal{P}_{sty}$ as style tokens $c_{sty}$. Simultaneously, a spatial conditioning image $I_{spt}$ is which is a composite of 3DMM landmarks and foreground masks, is computed from the target image $I_{tgt}$. Then, the denoising UNet, containing trainable low rank matrices, is optimized to disentangle identity and domain style, conditioned with $c_{id}$ and a ControlNet signal which combines $I_{spt}$ and $c_{sty}$.}
    \label{fig:model}
\end{figure*}

%% file: sec/3_method.tex
\section{Proposed Method}
\label{sec:method}
\subsection{Overview}
An overview of our proposed method is provided in Fig, \ref{fig:model}. We first adopt a \textit{training-free} style transfer technique to augment images from the VoxCeleb1 \cite{nagrani2020voxceleb} dataset. Next, we jointly fine-tune a diffusion model with LoRA \cite{hu2022lora}, along with two encoders that learn complementary representations for identity and style. This way the model can learn to accurately retarget faces irrespective of domain style.
\subsection{Injecting Domain Style for Data Augmentation}
\label{subsec:styleid}
Training on only real-world data leads to entanglement of identity and domain style leading to subpar retargeting. This can be ideally resolved by training with data from different sources \cite{ma2024follow, wei2024aniportrait}. However, such datasets are not readily available. Hence, we adopt a light diffusion-based style transfer technique to augment real-world data with different visual styles.\par
Chung \etal \cite{chung2024style} propose a self-attention-based style transfer method that fuses the style (texture) from a style image $I^S$ with the content (semantics and spatial layout) from a content image $I^C$. Their key observation is that \emph{queries} from content image can be matched with corresponding \emph{keys} in the style image when they share semantic similarities, thus maintaining spatial coherence. First, both images are inverted via DDIM inversion and the $(Q_t,K_t,V_t)$ are collected from each timestep $t$. As shown in Fig.\ref{fig:sty-inject},  queries from the  content image ($Q^c$), and key-value pairs $(K^s, V^s)$ from the style image are \emph{injected} into the decoder layers and are matched with the queries $Q^{cs}$. Specifically, they define the style injection operation at time step $t$ as follows:
\begin{align}
    \widetilde{Q}^{cs}_t &= \gamma \times Q^{c}_t + (1 - \gamma) \times Q^{cs}_t,\\
    \phi^{cs}_{\text{out}} &= \text{Attn}\bigl(\widetilde{Q}^{cs}_t, K^{s}_t, V^{s}_t\bigr)
\end{align}
where $Q^c_t$, $Q^{cs}_t$ are queries of the content and stylized images respectively and $K^s_t$, $V^s_t$ denote the key and value of the style image. $\gamma$ is a hyperparameter that determines the extent of content preservation. However, style injection alone is insufficient to harmonize the global appearance, because the initial noise in the diffusion process heavily influences color tones and structural cues. Denoising strictly from inverted content $z^c_T$ biases the output’s color, while denoising from inverted style $z^s_T$ biases its structure. To strike a balance, we employ Adaptive Instance Normalization (AdaIN) \cite{huang2017arbitrary} on both the content and style latents:
\begin{equation}
    z^{cs}_T = \sigma(z^s_T)\left(\frac{z^c_T\;-\;\mu(z^c_T)}{\sigma(z^c_T)}\right)\;+\;\mu(z^s_T)
\end{equation}
This operation aligns the distribution of content and style latents, mitigating unwanted color or structural shifts. \par
The entire stylization process just depends on DDIM inversion, so we use the pre-trained UNet $f_\theta$ to augment the images in the preprocessing step. We avoid direct integration with the training loop. This gives us two benefits (i) reduces complexity of the training, (ii) gives us the opportunity to filter out images with extreme artefacts by running a face detector model.
\subsection{Encoding Identity}
\label{subsec: method/id}
Identity embeddings from face recognition models (e.g., ArcFace) offer a compact and discriminative representation, but integrating them into existing diffusion models poses a challenge. Arc2Face \cite{papantoniou2024arc2face} addresses this by retraining the CLIP text encoder to interpret ID embeddings wrapped in a pseudo-token, effectively enforcing identity fidelity. However, this approach limits the influence of other control signals such as pose or style, leading to rigid retargeting behaviour and poor style transfer, as seen in our experiments. To balance identity fidelity with controllability, we propose a shallow transformer-based module $\mathcal{P}_{id}$ that maps identity embeddings $f_{id}\in \mathbf{R}^{512}$ into the CLIP text space. This allows the model to preserve identity while remaining responsive to additional conditioning signals. The identity conditioning $c_{id}$ can be formulated as:
\begin{align}
    c_{id} = \mathcal{P}_{id} (f_{id}) = \mathcal{P}_{id} (\mathcal{E}_{id}(I_{src}))
\end{align}
\subsection{Incorporating Spatial Control}
\label{method/spatial}
For spatial control, we first render the target facial landmarks using a parametric face model. To mitigate background artifacts during synthesis, we blend the foreground masks of the source and target images, ensuring cleaner identity transfer. The resulting composite is then used as a conditioning signal for the denoising process via a ControlNet module \cite{deng2019arcface}, allowing precise guidance of facial geometry during generation.
\subsection{Encoding Domain Style}
\label{method/style}
While identity embeddings ensure discriminative control, they lack the rich perceptual detail required for realistic appearance transfer. CLIP embeddings, as observed in prior works like IP-Adapter \cite{ye2023ip}, emphasize perceptual likeness- capturing cues such as lighting, hair, and local texture. To leverage this, we extract visual embeddings from the source image using the CLIP encoder $\mathcal{E}_{sty}$, leveraging both patch-level and global tokens. They are projected into the text embedding space via a shallow transformer decoder $\mathcal{P}_{sty}$ with learnable queries, mirroring our identity encoder. The resulting tokens $c_{sty}$ supplement the identity tokens $c_{id}$, enabling more faithful preservation of domain-specific details during retargeting.
\begin{align}
    c_{sty} = \mathcal{P}_{sty} (\mathcal{E}_{sty} (I_{src}))
\end{align}
\subsection{Optimization}
\label{subsec:finetune}

The identity tokens $c_{id}$ are fed to the denoising UNet directly and the style tokens $c_{sty}$ are fused with spatial condition $c_s$ in the ControlNet to produce the conditioning signal $c' = \mathcal{C} (c_{s}, c_{sty})$. This is fed to the UNet decoder via zero convolutions.\par
We jointly train the transformer modules of the identity and style encoders, along with the ControlNet branch for spatial conditioning. Additionally, we observe that introducing low-rank adaptation (LoRA) layers on top of the frozen denoising U-Net consistently improves generation quality compared to training without them (refer to ablation experiments in Section\ref{sec:abl}).
Formally, the training objective can be described as:
\begin{align}
   \mathcal{L} = ||\hat{\epsilon_t} - \epsilon ||_2^2 =  ||f_{\theta +\Delta\theta}(x_t, c_{id}, c') - \epsilon||_2^2
\end{align}
where, $f_{\theta +\Delta\theta}$ represents the denoising UNet, with frozen parameters $\theta$ and trainable LoRA parameters $\Delta\theta$. 
\subsection{Implementation Details}
\textbf{Architecture}: The transformer modules $\mathcal{P}_{id}$ and $\mathcal{P}_{sty}$ used to map the identity and style embeddings into CLIP textual space have three-layers, following the implementation of Han \etal \cite{han2024face}. We use ArcFace \cite{deng2019arcface} embeddings to represent identity. For obtaining facial landmarks, we use the Deep3DFaceRecon \cite{deng2019accurate} model, and couple it with an off-the-shelf segmentation network for foreground mask extraction. We use Stable Diffusion v1-5 as our base UNet. \\
\textbf{Training Hyperparameters}: We train our model for 40,000 epochs with a constant learning rate of 1e-4 and a batch size of 4 on a 4x NVIDIA A5000 GPU setup.

%% file: sec/4_experiments.tex
\section{Experiments}
\label{sec:results}
\subsection{Datasets}
For evaluation, we choose 20 subjects from VoxCeleb1 test split, each with 3 video sequences. We extract 10 frames from each, giving us a total of 600 frames. For augmentation we choose from a set of 5 domain styles, with varying degrees of abstraction, some of which are shown in Fig. \ref{fig:augment}. This gives us a total of 3000 augmented frames which are used for quantitative analysis (reported in Tables \ref{tab:quant_main} and \ref{tab:ablation}). All models have been evaluated in self and cross-identity settings. Further comparisons are made using 3 subjects from the GeneFace \cite{ye2023geneface} dataset (see Supp \ref{sec:supp_addn}). We also use in-the-wild faces and domains styles to gauge the generalization abilities of our method.
\input{sec/figures/augment.tex}
\subsection{Evaluation Metrics}
\textbf{Self Retargeting}: \textbf{PSNR} and \textbf{LPIPS} is used to measure reconstruction quality. We use cosine similarity between identity embeddings (dubbed as \textbf{CS-ID}) to measure identity retention, as previously done in \cite{bounareli2023hyperreenact, han2024face}. For expression and pose, we measure \textbf{motion transfer error}, given by the Euclidean distances between the expression and pose coefficients of the generated and driving images. To measure stylistic similarity between the generated and ground truth images, we use a recently proposed \textbf{ArtFID} \cite{wright2022artfid} metric which evaluates both content and style preservation and strongly coincides with human judgement. It is calculated as ArtFID = (1+LPIPS)(1+FID), where LPIPS measures content similarity between the generated image and original ground truth image, and FID measures the style similarity between the generated image and augmented ground truth image.\\
\textbf{Cross Identity Retargeting}: We measure identity similarity and retargeting fidelity using the same metrics as in self retargeting. Since ground truth images are not available in this setting we cannot employ reconstruction-based metrics.

\input{sec/figures/styleinject.tex}
\subsection{Comparisons with Existing Methods}
Although there are no image-based models that directly address cross-domain face retargeting, we compare against existing state-of-the-art face synthesis models based on GANs, diffusion and neural textures: \textbf{HyperReenact} \cite{bounareli2023hyperreenact} learns a hyper-network that offsets the weights of a frozen StyleGAN generator to achieve target pose. \textbf{DiffusionRig} \cite{ding2023diffusionrig} conditions a diffusion model with 3DMM renderings, akin to GIF \cite{GIF2020}, but improves photometric reconstruction using an universal appearance encoder. It also uses a separate fine-tuning stage to adapt the model to individual subjects at test time with a small album. \textbf{Arc2Face} \cite{papantoniou2024arc2face} is the state-of-the-art identity-aware diffusion model conditioned solely on ArcFace \cite{deng2019arcface} embeddings. Similar to ours, it enables retargeting with a ControlNet conditioned on 3DMM renderings. \textbf{ROME} \cite{khakhulin2022realistic} uses neural textures for one-shot face retargeting in the wild. By using these baselines, we provide sufficient comparisons on multiple metrics. Note that due to compute constraints, video diffusion models are not considered. We provide more details on the differences in training datasets of the baselines in the Supp. \ref{sec:supp_baseline}.
\subsubsection{Qualitative Results}
Fig. \ref{fig:qual1} shows a visual comparison among various models for cross-domain face retargeting. We observe that \textbf{\textit{StyleYourSmile}} outperforms existing ones in self and cross-identity settings. Arc2Face \cite{papantoniou2024arc2face} is primarily trained on identity embeddings and spatial control is proposed as an add-on. As a result, it struggles to constrain the face to the underlying mesh and completely overlooks domain-specific cues. DiffusionRig \cite{ding2023diffusionrig} requires a personal album for inference-time fine-tuning. As our task is defined in a single image-to-image setting, it falls short of it's full potential. However, it is able to capture the high-level details of the source image like face orientation and overall colour tone. For the sake of completeness, we make additional comparisons where we fine-tune it on small image sets. Please refer to Supp. \ref{ref:supp_diffrig} for more details. HyperReenact \cite{bounareli2023hyperreenact} shows good retargeting fidelity but is unable to capture domain-specific details and fine-grained facial features accurately. ROME \cite{khakhulin2022realistic} shows competitive retargeting performance when the foreground is accurately segmented. However, this cannot be guaranteed in in-the-wild images. It also fails to capture texture details correctly.
\input{sec/figures/main_results.tex}
\input{sec/tables/main_table.tex}

\subsubsection{Quantitative Results}
We report the performance of all models on different metrics in Table \ref{tab:quant_main}. In self retargeting setting, our model achieves highest reconstruction quality and style preservation. Although, Arc2Face \cite{papantoniou2024arc2face} generates visually disparate results compared to the source domain, it achieves high scores on CS-ID metric, demonstrating their ability to preserve identity in challenging scenarios. This can be also attributed to the fact that it was explicitly trained for identity retention with face recognition embeddings.

\subsection{Out-of-Domain Performance}
We observe that our model can generalize to domains unseen in the training set. As shown in Figure \ref{fig:ood}, we augment in-the-wild portraits with unseen domain styles and yet our model is able to faithfully retarget expressions without compromising subject identity or domain-specific attributes of the source image.
\input{sec/figures/ood.tex}

%% file: sec/figures/augment.tex
\begin{figure}
    \centering
    \includegraphics[width=0.7\linewidth]{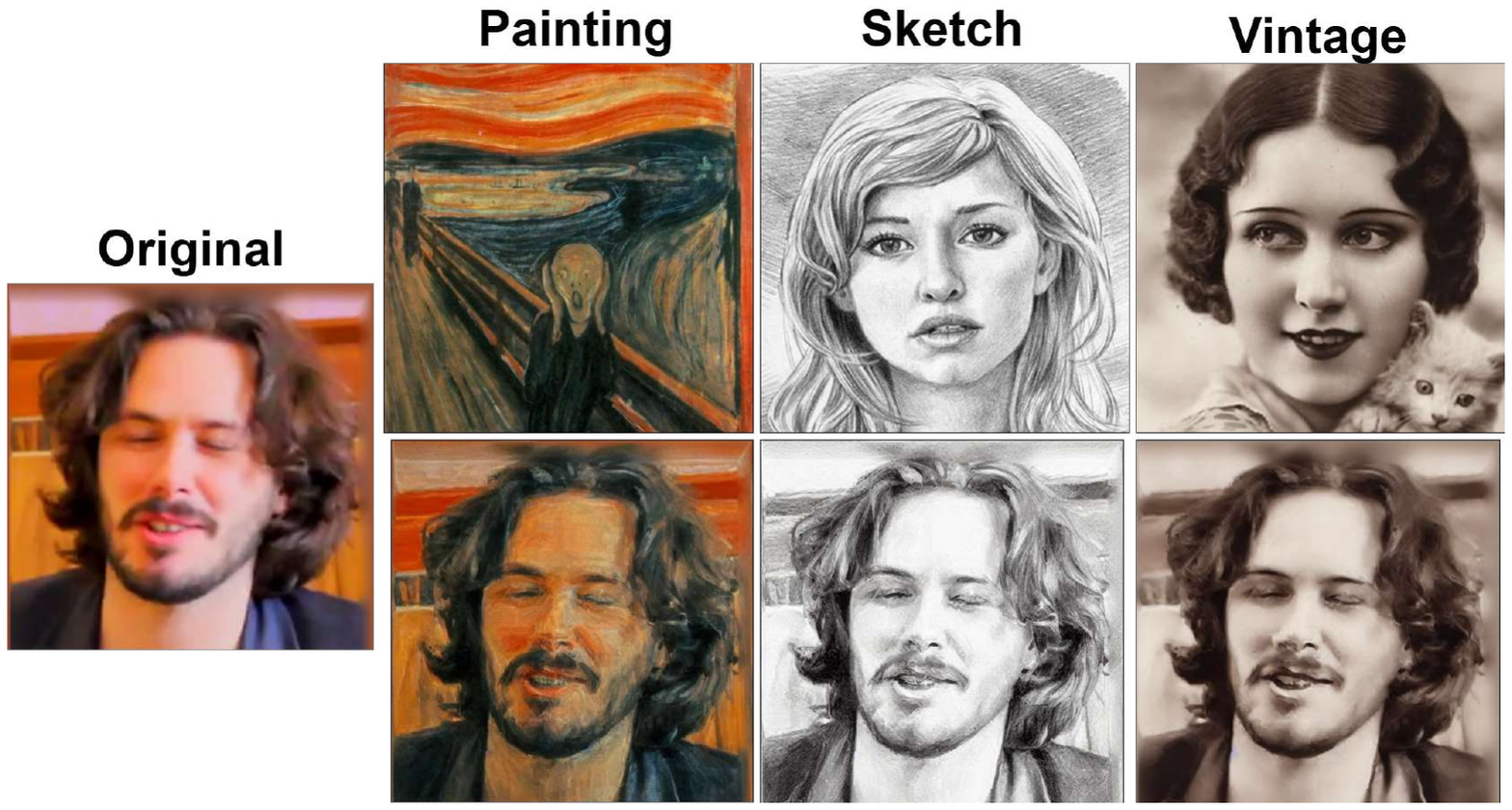}
    \caption{We augment the training data with different styles, with varying degrees of abstraction. Training on such data incentivize the model to decouple identity from image style.}
    \label{fig:augment}
\end{figure}

%% file: sec/figures/styleinject.tex
\begin{figure}
    \centering
    \includegraphics[width=\linewidth]{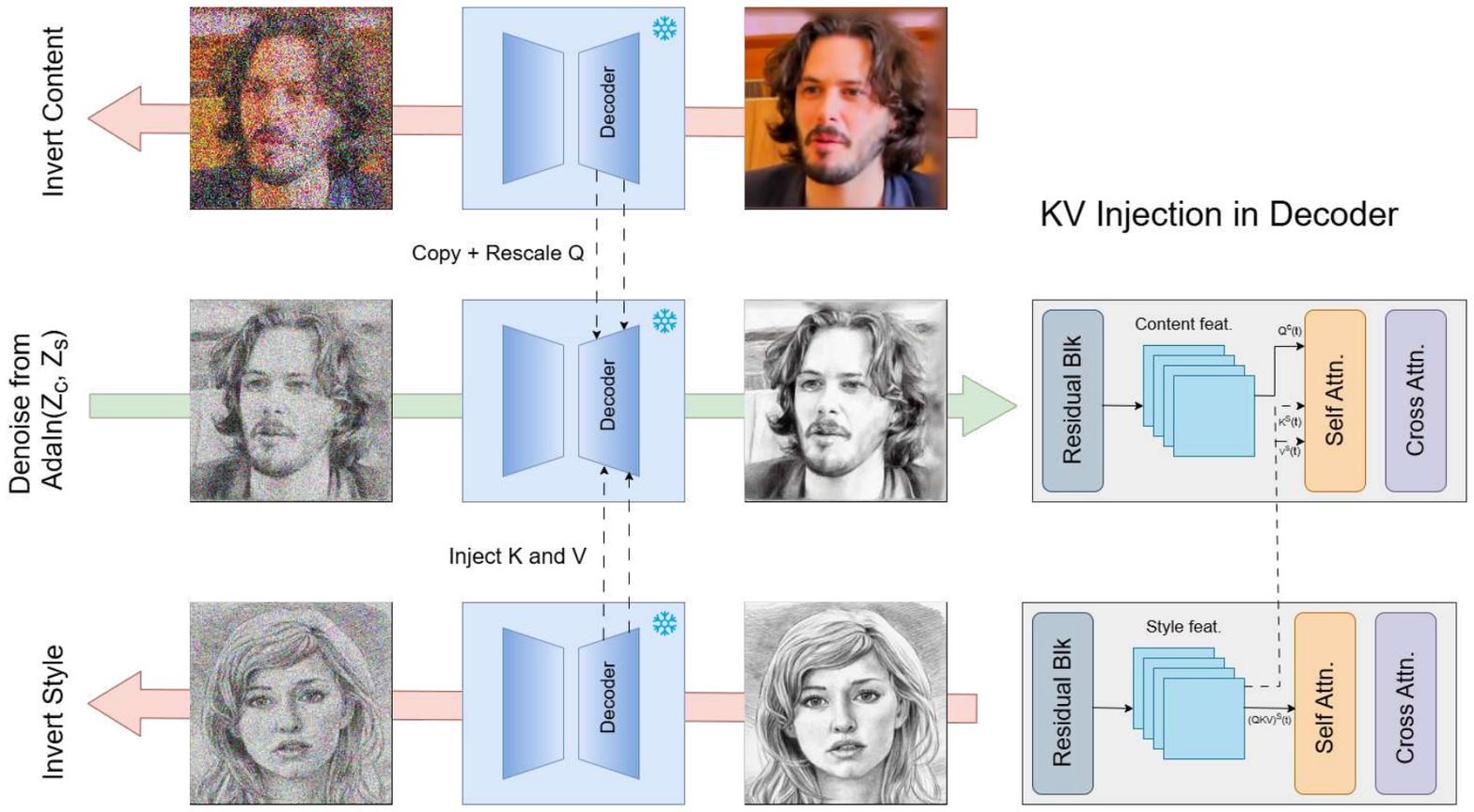}
    \caption{Style Injection Method \cite{chung2024style}: First, the content and style images are inverted into latents $z^c_T$ and $z^s_T$ respectively. During inversion ($Q,K,V$) of both are cached. For generating the styled image, we start with AdaIn($z^c_T$,$z^s_T$) and inject the key-value pairs ($K^s,V^s$) from the style image into the decoder layers, where they are matched with corresponding queries $\Tilde{Q}^{cs}$}
    \label{fig:sty-inject}
\end{figure}

%% file: sec/figures/main_results.tex
\begin{figure*}
    \centering
    \includegraphics[width=\linewidth]{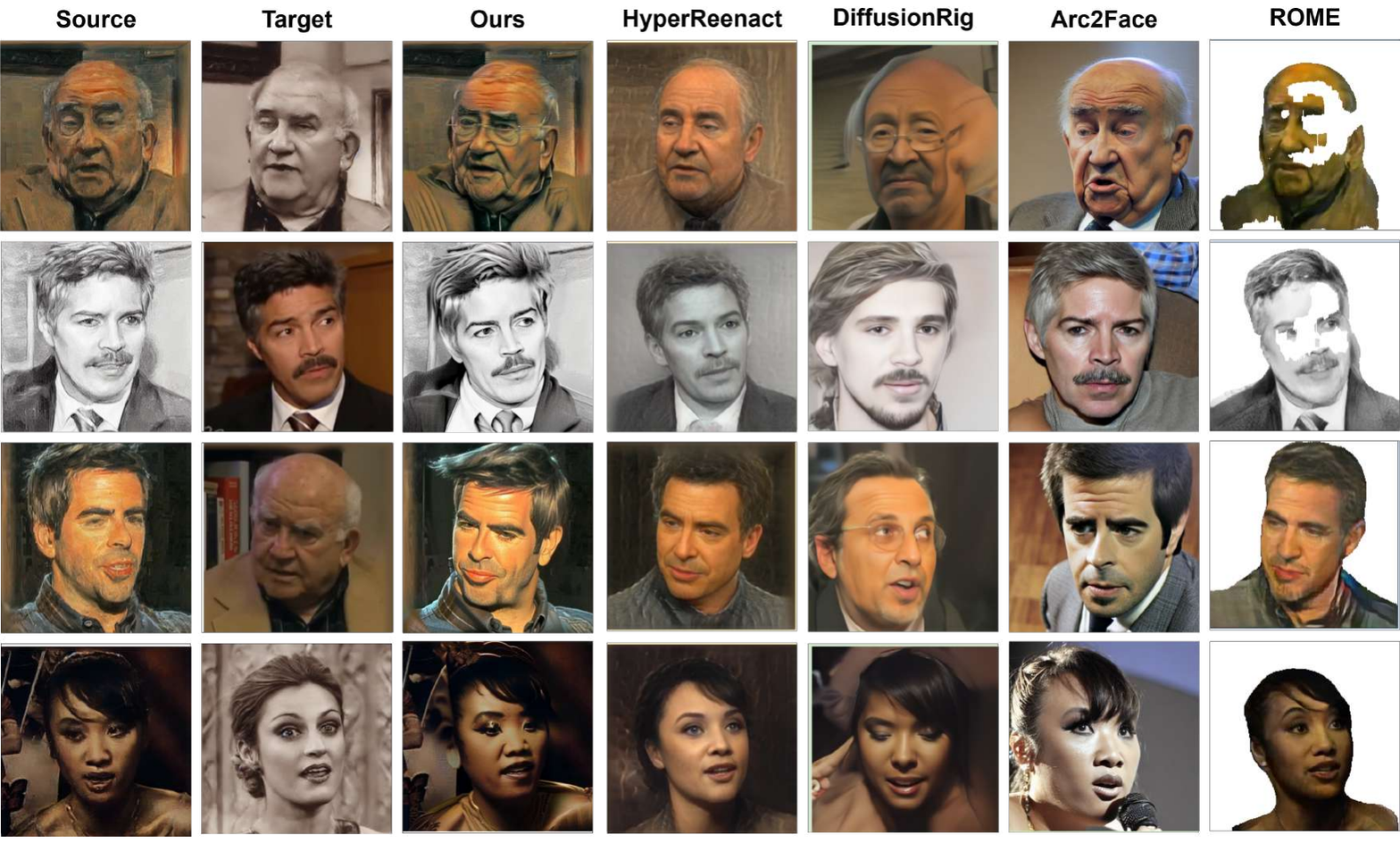}
    \caption{Visual comparison of various models on stylized VoxCeleb1  \cite{nagrani2020voxceleb} test set. Our model outperforms previous models in terms of identity retention and style preservation.}
    \label{fig:qual1}
\end{figure*}

%% file: sec/tables/main_table.tex
\begin{table*}[t!]
\centering
\begin{tabularx}{\textwidth}{l|XXXXXX|XXX}
\toprule
\multirow{2}{*}{\textbf{Methods}} & \multicolumn{6}{c|}{\textbf{Self Retargeting}} & \multicolumn{3}{c}{\textbf{Cross ID Retargeting}} \\
\cline{2-10}
 & \multicolumn{1}{c|}{PSNR$\uparrow$} & \multicolumn{1}{c|}{LPIPS$\downarrow$} & \multicolumn{1}{c|}{CS-ID$\uparrow$}  & \multicolumn{1}{c|}{Exp$\downarrow$} & \multicolumn{1}{c|}{Pose$\downarrow$} & \multicolumn{1}{c|}{ArtFID$\downarrow$} & \multicolumn{1}{c|}{CS-ID$\uparrow$} & \multicolumn{1}{c|}{Exp$\downarrow$} & \multicolumn{1}{c}{Pose$\downarrow$} \\
\midrule
HyperReenact \cite{bounareli2023hyperreenact} & 12.225 & 0.377 & 0.410 & 0.368 & 7.334 & 35.536 & 0.270 & 0.387 & \textbf{6.344} \\
ROME \cite{khakhulin2022realistic} & 10.037 & 0.511 & 0.189 & 0.491 & 8.486 & 38.002 & 0.091  & 0.420 & 8.375\\
Arc2Face \cite{papantoniou2024arc2face} & 9.403 & 0.455 & \textbf{0.635} & 0.394 & 7.198 & 41.177 & \textbf{0.606} & 0.551 & 9.071 \\
DiffusionRig \cite{ding2023diffusionrig}& 13.650 & 0.402 & 0.324 & 0.273 & 7.034 & 35.392 & 0.221 & 0.414  & 9.111\\
Ours & \textbf{19.889} & \textbf{0.146} & \underline{0.615} & \textbf{0.241} & \textbf{6.321} & \textbf{32.377} & \underline{0.553} & \textbf{0.333} & 8.072 \\
\bottomrule
\end{tabularx}
\caption{Quantitative evaluations among various methods for cross-domain face retargeting on Voxceleb1 test set. Best figures are given in \textbf{bold} and second-best figures are \underline{underlined}.}
\label{tab:quant_main}
\end{table*}

%% file: sec/figures/ood.tex
\begin{figure}
    \centering
    \includegraphics[width=0.8\linewidth]{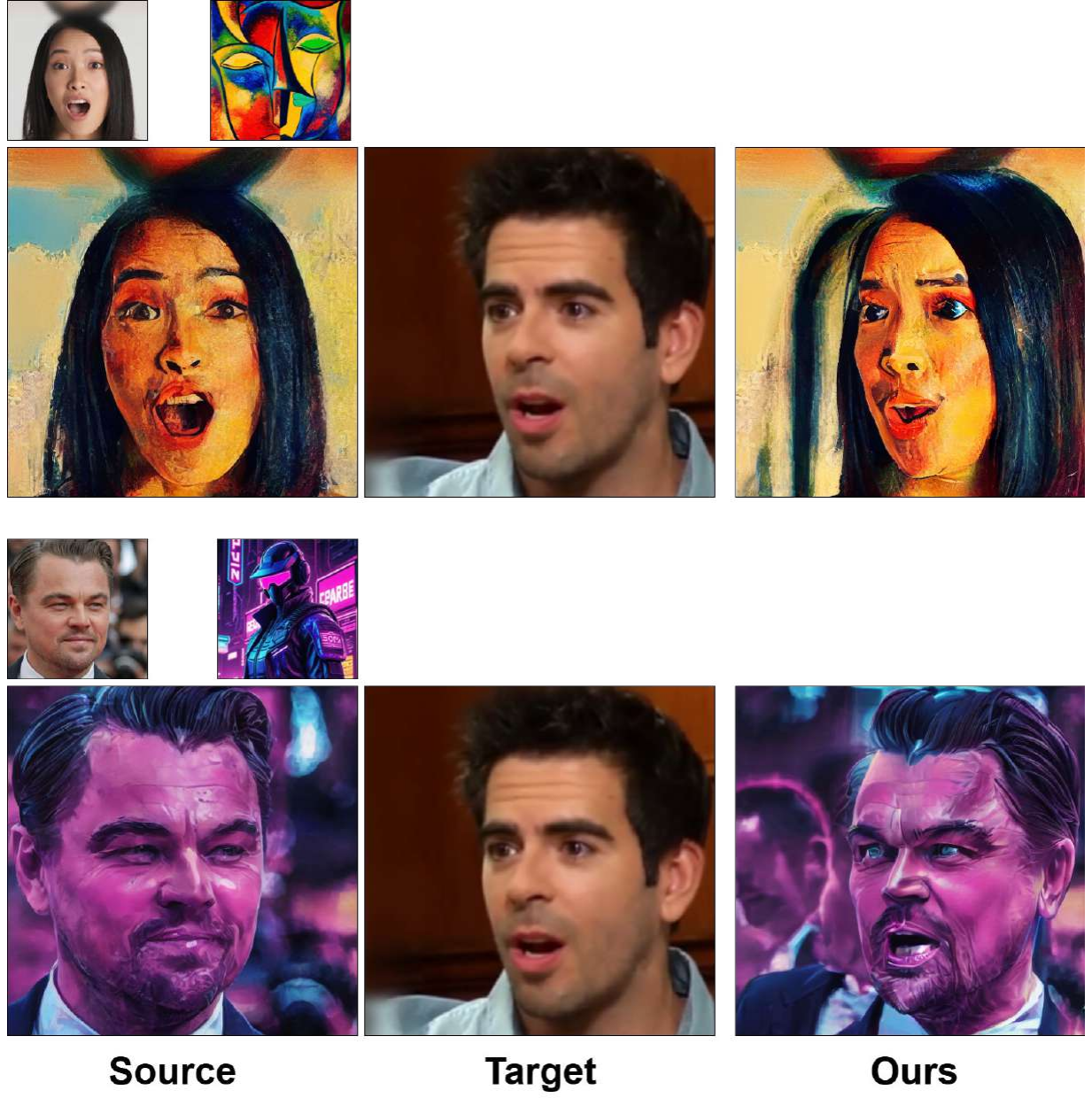}
    \caption{OOD Generalization of our model: We
have picked two in-the-wild portraits and two unseen styles. As
shown, our model is able to preserve the domain-specific cues and subject identity accurately.}
    \label{fig:ood}
\end{figure}

%% file: sec/5_ablations.tex
\section{Ablations}
\label{sec:abl}
\begin{itemize}
    \item \textbf{\textit{Style Augmentation}}: The style augmentation module determines the quality of the training data, which impacts the final model performance. We experiment with different models as shown in  Table \ref{tab:style_transfer}. StyleInject \cite{chung2024style} yields the optimal ArtFID score, implying best balance of content and style preservation. Furthermore, as shown in Fig. \ref{fig:gamma_abl} our choice of injection strength $\gamma$ gives the best ArtFID score.

    \item \textbf{\textit{Style Conditioning}}: Directly feeding the style features from CLIP to the denoising UNet distorts some facial features and the overall colour tone, as shown in Fig. \ref{fig:abl_fig} (C1). Routing the style features via the ControlNet module enables the UNet to preserve fine-grained identity features as well as the colour grading.

    \item{\textbf{\textit{Joint Fine-tuning}}}: Although a frozen UNet keeps the framework modular \cite{han2024face}, it inhibits the domain specific cues, as shown in Fig.\ref{fig:abl_fig} (C2). Hence we adopt a \textit{light-touch} optimization with LoRA which improves style preservation.
\end{itemize} 
\input{sec/figures/reb_gamma.tex}
\input{sec/tables/ablation_tab.tex}
\input{sec/tables/style_transfer.tex}
\input{sec/figures/ablation.tex}

%% file: sec/figures/reb_gamma.tex
\begin{figure}
    \centering
    \includegraphics[width=0.9\linewidth]{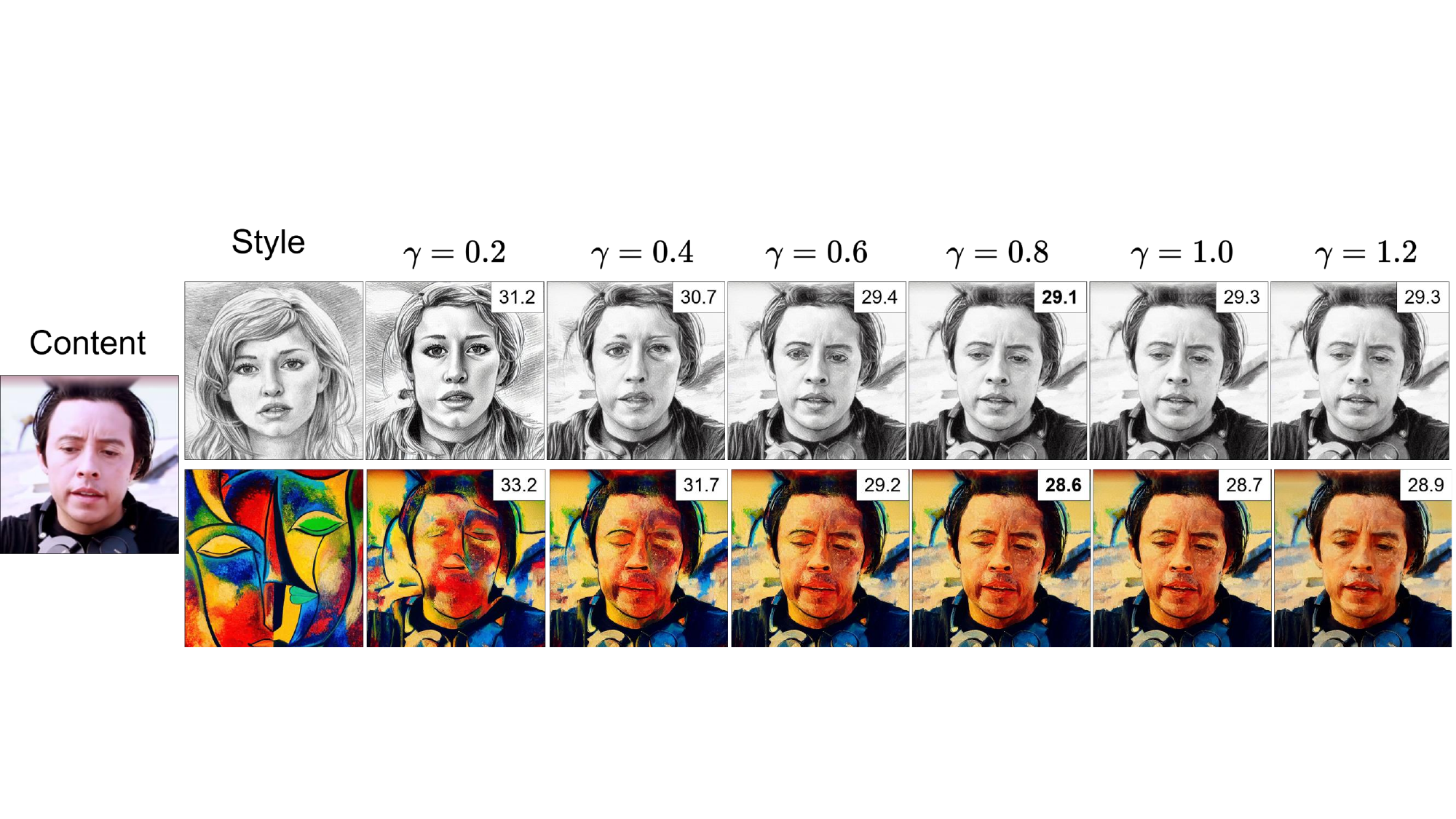}
    \vspace{-0.5cm}
    \caption{Effect of Style Injection Strength $\gamma$. Our choice of $\gamma=0.8$ yields optimal ArtFID score implying it offers the best balance between content and style.}
    \label{fig:gamma_abl}
\end{figure}

%% file: sec/tables/ablation_tab.tex
\begin{table}[]
\centering
\begin{tabular}{c|cc}
\hline
\multirow{2}{*}{\textbf{Config}}  & \multicolumn{2}{c}{\textbf{Metrics}} \\ \cline{2-3} 
                                  & \multicolumn{1}{c|}{ID$\uparrow$}     & Style$\downarrow$  \\ \hline
Config. C1                         & \multicolumn{1}{c|}{0.599}  & 32.414  \\
Config. C2          & \multicolumn{1}{c|}{0.607}  & 32.381  \\
Ours & \multicolumn{1}{c|}{\textbf{0.615}}  & \textbf{32.377 } \\ \hline
\end{tabular}
\caption{Ablation Experiments: We study the effect of (C1) feeding the style tokens directly to the UNet and (C2) joint fine-tuning. Best values are highlighted.}
\label{tab:ablation}
\end{table}

%% file: sec/tables/style_transfer.tex
\begin{table}[]
\begin{tabular}{c|cc}
\hline
\multicolumn{1}{l|}{\textbf{Method}} & \multicolumn{1}{l}{\textbf{ArtFID}} & \multicolumn{1}{l}{\textbf{Time (s)}} \\ \hline
\textbf{AdaIN \cite{huang2017arbitrary}}                       & 30.93                               & \textbf{0.06}                         \\
\textbf{AdaAttn \cite{liu2021adaattn}}                     & 30.35                               & 0.1                                   \\
\textbf{AdaConv \cite{chandran2021adaptive}}                     & 31.86                               & 0.08                                  \\
\textbf{DiffStyle \cite{jeong2023training}}                   & 41.46                               & 350                                   \\
\textbf{StyleInject (Ours)}                 & \textbf{28.08}                      & 5                                     \\ \hline
\end{tabular}
\caption{Comparison of various state-of-the-art style transfer methods. We choose StyleInject \cite{chung2024style} since it provides the optimal balance between content and style fidelity within a reasonable time budget.}
\label{tab:style_transfer}
\end{table}

%% file: sec/figures/ablation.tex
\begin{figure}
    \centering
    \includegraphics[width=0.8\linewidth]{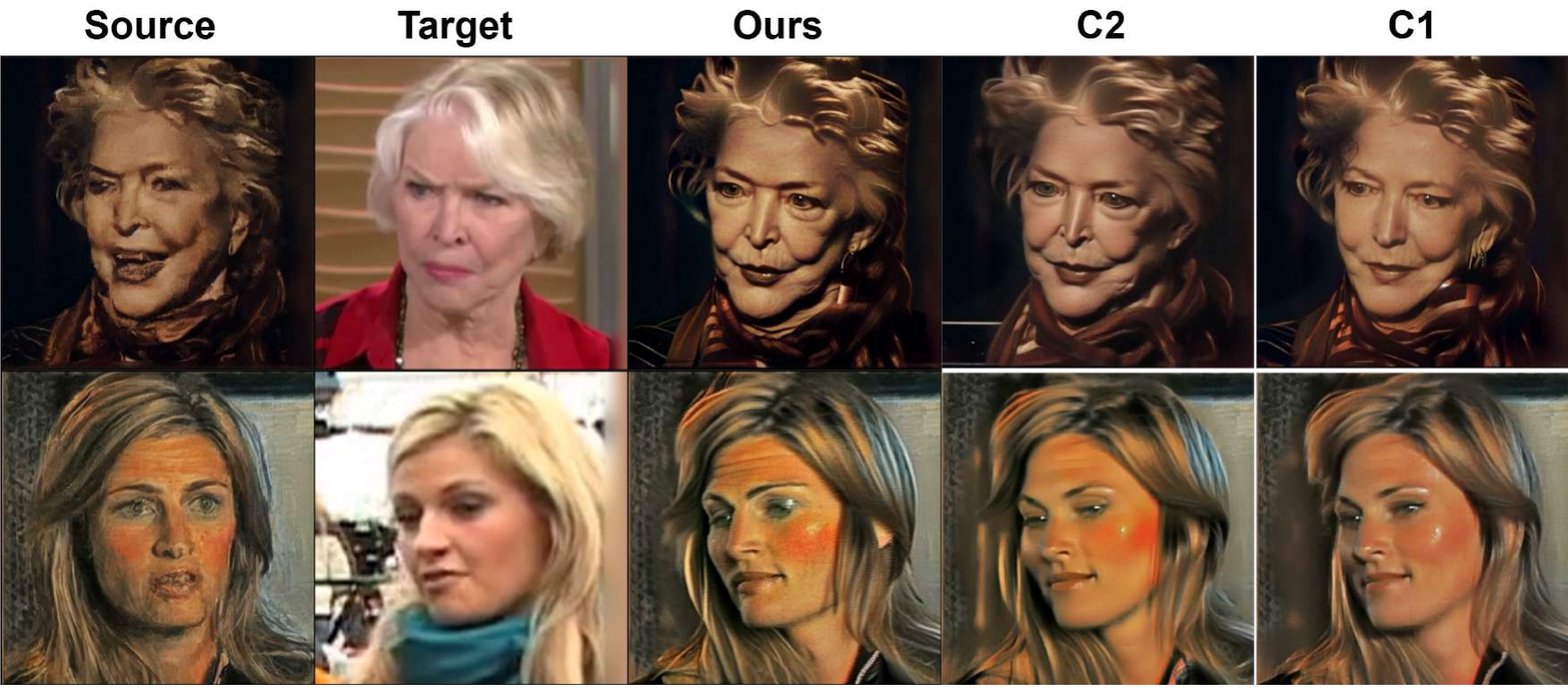}
    \caption{Ablation Experiments: Results showing the effect of (C1) feeding the style tokens $c_{sty}$ cirectly to the denoising UNet and (C2) keeping the denoising UNet frozen during training. Metrics reported in Table \ref{tab:ablation}}
    \label{fig:abl_fig}
\end{figure}

%% file: sec/6_conclusion.tex
\vspace{-0.4cm}
\section{Social Impact}
While our model has practical value for entertainment purposes, the potential misuse of it can lead to misinformation, and ethical concerns. To prevent such cases, digital watermarks can be used to determine  authenticity of the content.
\vspace{-0.4cm}
\section{Conclusion}
In this work, we introduce \textbf{\textit{StyleYourSmile}}, an unified framework to solve cross-domain face retargeting with diffusion models. We hypothesize that existing methods have entangled representations for identity and style since they are trained on a singular domain. First, we adopt a light training-free image stylization method that augments video frames with abstract styles from multiple domains. Then, we train two complementary encoders to learn disentangled representations for identity and domain style. To encourage better style preservation, we jointly optimize the denoising UNet with LoRA. Through extensive evaluations, we demonstrate that our model outperforms existing ones in terms of identity retention and style preservation.\par

%% file: sec/suppl.tex
\clearpage
\setcounter{page}{1}
\maketitlesupplementary

\section{Limitations and Future Work}
We evaluate our model on challenging portraits with occlusions, CGI effects and higher abstractions (like anime). Fig. \ref{fig:wild_suppl} shows some failure cases of our model. Firstly, it completely fails in case of animations as the underlying face detector does not recognize such faces. In the third row, we see that the model ignores the eye patch, which is relevant to the source image, but does transfer the expressions correctly. Our model suffers from significant identity drift when faces have complex CGI effects. Future work can address these problems. While large video diffusion models can handle such cases, we encourage the development of more efficient solutions to diffusion-based retargeting.
\section{Additional Comparisons with DiffusionRig}
\label{ref:supp_diffrig}
In our main experiments, we compare DiffusionRig \cite{ding2023diffusionrig} without Stage 2 i.e. no personalization on personal album, since our task is in a single frame setting. However, for the sake of completeness, we conduct additional comparisons with a fine-tuned version. We select 10 frames that are allied to the source image and fine-tune the model for 1000 epochs. This takes around 10 minutes on a NVIDIA A5000 GPU. We observe that this leads to memorization and the model ignores the target landmarks altogether. This problem persists regardless of the dataset. Fig. \ref{fig:macron} shows a similar phenomenon when performing self retargeting on GeneFace \cite{ye2023geneface} videos. We hypothesize that this is correlated with the diversity of the personal album since in the official implementation the authors have used portraits collected from different sources. Validating this idea, is beyond the scope of this paper but this observation paves the way for further analysis.
\input{sec/figures/diffrig_supp1.tex}
\input{sec/figures/diffrig_supp2.tex}
\input{sec/figures/limitation.tex}
\section{Additional Details on Existing Baselines}
\label{sec:supp_baseline}
Our model, \textit{\textbf{StyleYourSmile}}, outperforms several baselines for face retargeting but it is necessary to highlight the differences in training processes among them for better judgement. HyperReenact \cite{bounareli2023hyperreenact} builds on StyleGAN2 \cite{karras2020analyzing} and is trained on VoxCeleb1 \cite{nagrani2020voxceleb}, while ROME \cite{khakhulin2022realistic} relies on VoxCeleb1 \cite{nagrani2020voxceleb} and VoxCeleb2 \cite{chung2018voxceleb2}, ensuring fair comparisons. Arc2Face uses large-scale datasets—LAION Faces \cite{zheng2022general} and Web260M \cite{zhu2021webface260m}, respectively—with over 50M images, facilitating robust representations. DiffusionRig \cite{ding2023diffusionrig}, however, is trained on the FFHQ dataset \cite{karras2019style} ($\approx$70K images). DiffusionRig also requires atleast 8× A100 GPUs for retraining, and assumes a personal album for test-time fine-tuning, rendering its comparison less equitable. Nonetheless, we include all relevant baselines to provide a comprehensive view of the problem space.
\section{Additional Results for Face Retargeting}
\label{sec:supp_addn}
In Fig. \ref{fig:supp_geneface} and Fig. \ref{fig:supp_qual1} we show additional qualitative results of on GeneFace \cite{ye2023geneface} and VoxCeleb1 \cite{nagrani2020voxceleb} images. Our model achieves a better balance of identity, style and reenactment fidelity in challenging cross-domain settings.
\input{sec/figures/geneface_supp.tex}
\input{sec/figures/result_supp1.tex}

%% file: sec/figures/diffrig_supp1.tex
\begin{figure}
    \centering
    \includegraphics[width=0.7\linewidth]{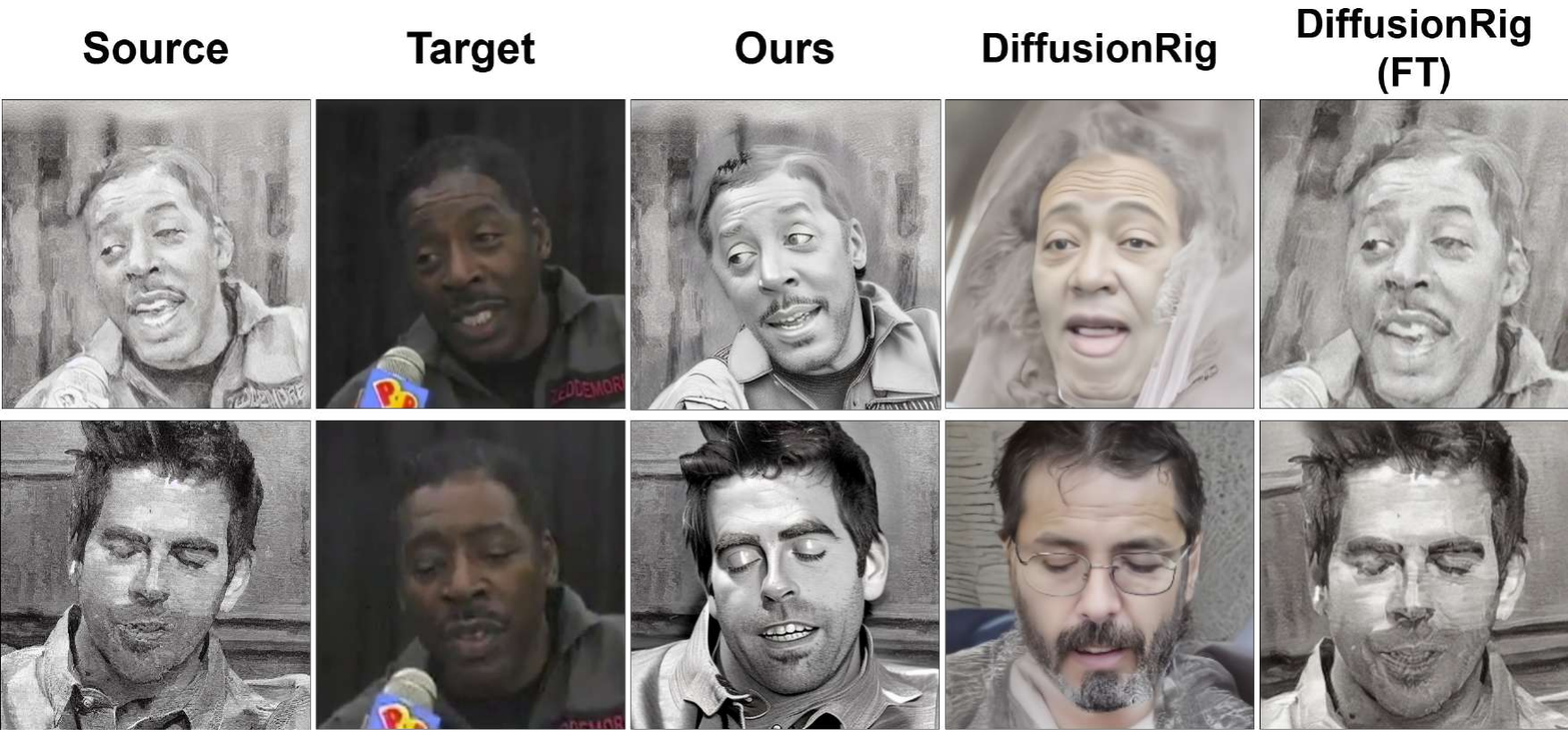}
    \caption{Additional Comparison with DiffusionRig. Without fine-tuning, DiffusionRig fails to learn the identity correctly. But using allied frames from the same video as the source leads to memorization.}
    \label{fig:enter-label}
\end{figure}

%% file: sec/figures/diffrig_supp2.tex
\begin{figure}
    \centering
    \includegraphics[width=\linewidth]{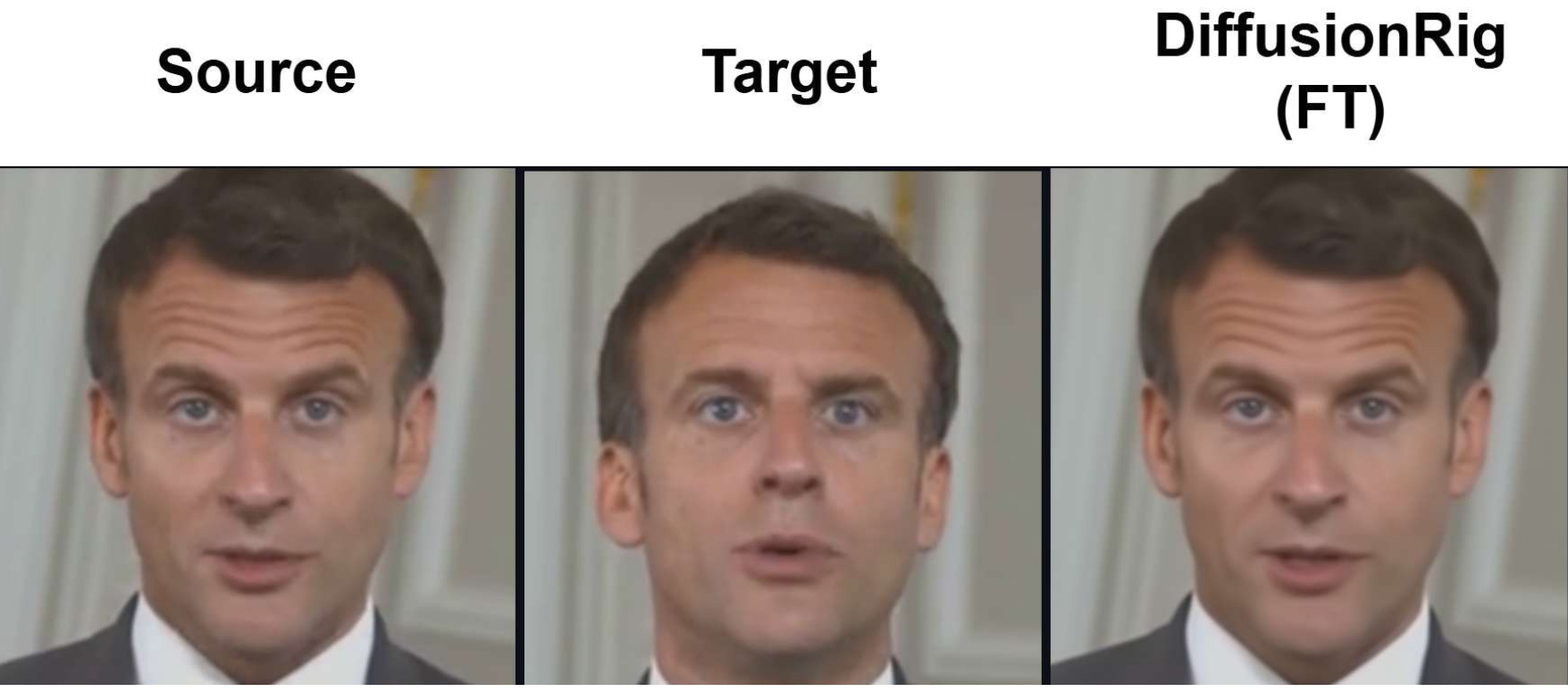}
    \caption{Fine-tuning DiffusionRig on perceptually similar images leads to memorization and the model ignores the conditioning signal altogether.}
    \label{fig:macron}
\end{figure}

%% file: sec/figures/limitation.tex
\begin{figure}
    \centering
    \includegraphics[width=0.8\linewidth]{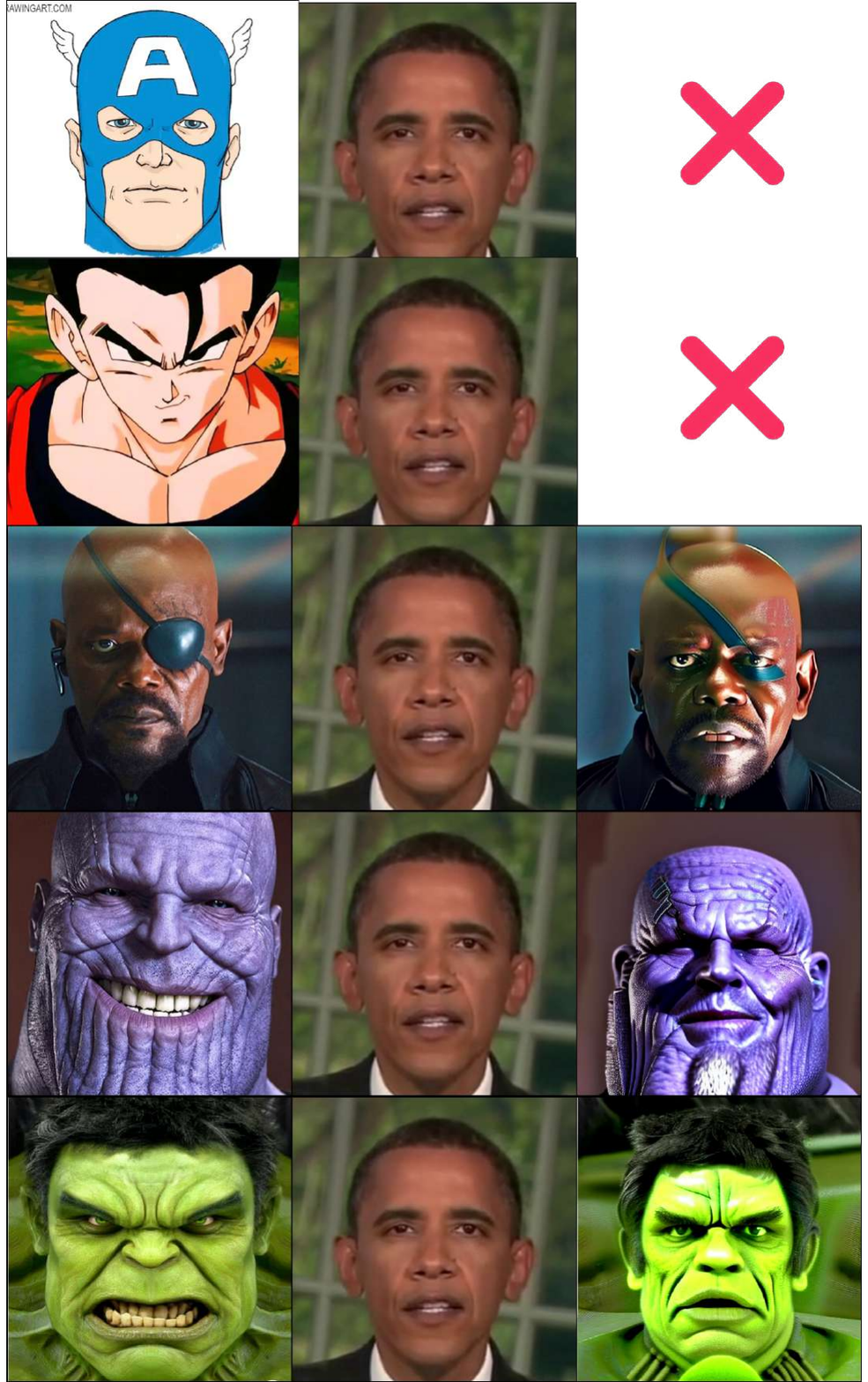}
    \caption{We assess our model's performance when the source image is \textbf{not spatially correlated} to the training data. Here we take faces with occlusions, complex CGI effects and anime. }
    \label{fig:wild_suppl}
\end{figure}

%% file: sec/figures/geneface_supp.tex
\begin{figure*}
    \centering
    \includegraphics[width=\linewidth]{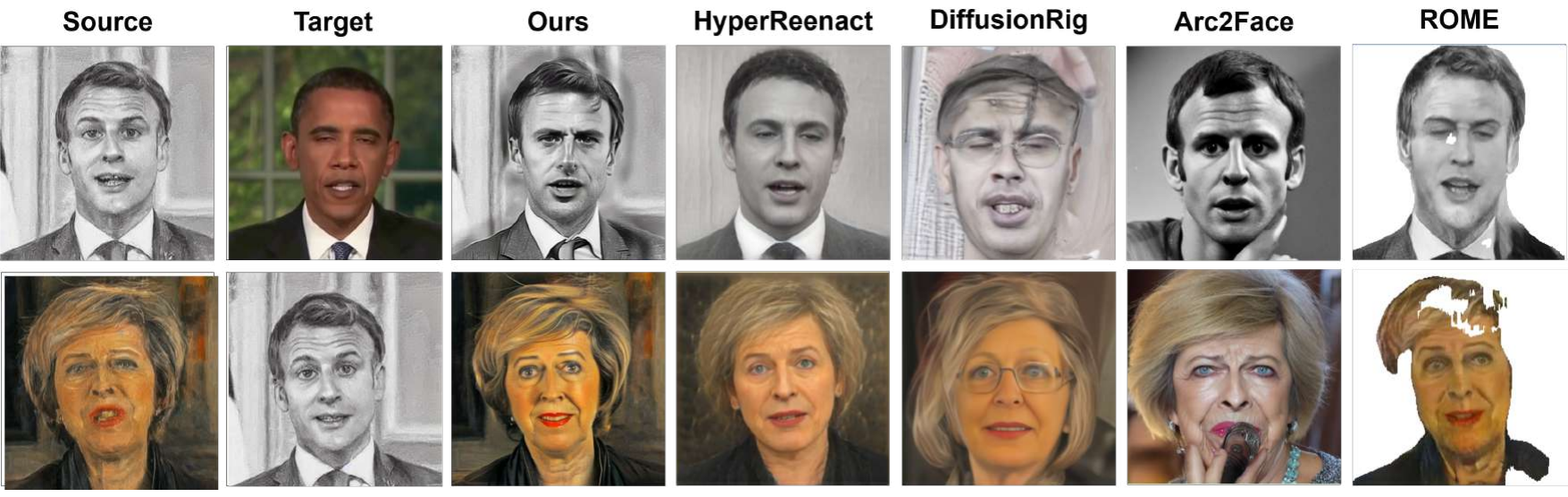}
    \caption{Visual comparison on GeneFace \cite{ye2023geneface} data. Our model outperforms other models in terms of identity retention and retargeting fidelity.}
    \label{fig:supp_geneface}
\end{figure*}

%% file: sec/figures/result_supp1.tex
\begin{figure*}
    \centering
    \includegraphics[width=\linewidth]{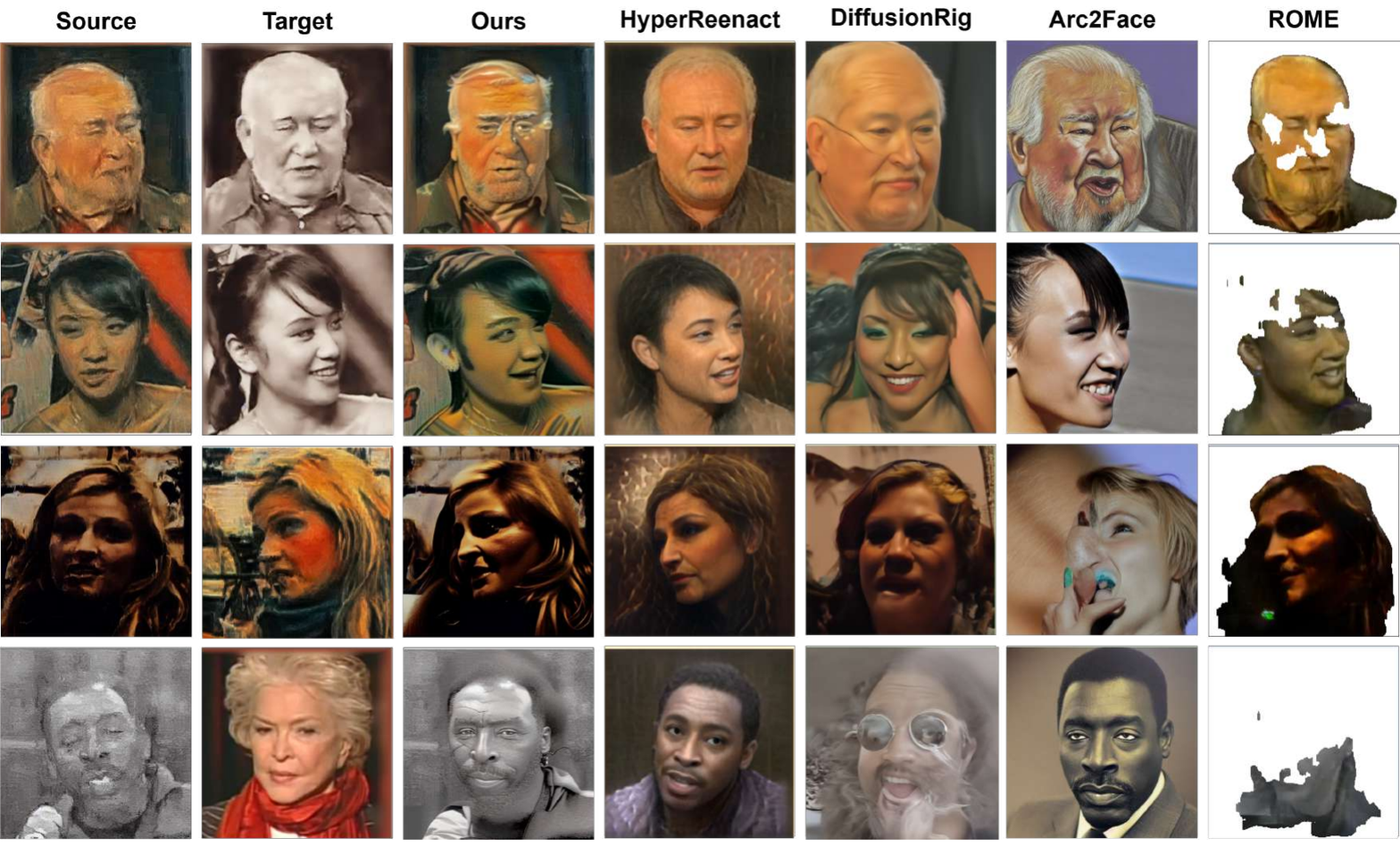}
    \caption{Visual comparison of various models on VoxCeleb1 \cite{nagrani2020voxceleb} test set. Our model outperforms other models in terms of identity retention and retargeting fidelity.}
    \label{fig:supp_qual1}
\end{figure*}

%% file: main.bib
@String(ICCV= {Int. Conf. Comput. Vis.})

@String(TOG= {ACM Trans. Graph.})

@String(ICLR = {Int. Conf. Learn. Represent.})

@String(AAAI = {AAAI})

@String(ICCV  = {ICCV})

@String(TOG   = {ACM TOG})

@String(ICLR  = {ICLR})

@inproceedings{karras2020analyzing,
  title={Analyzing and improving the image quality of stylegan},
  author={Karras, Tero and Laine, Samuli and Aittala, Miika and Hellsten, Janne and Lehtinen, Jaakko and Aila, Timo},
  booktitle={Proceedings of the IEEE/CVF conference on computer vision and pattern recognition},
  pages={8110--8119},
  year={2020}
}

@inproceedings{GIF2020,
    title = {{GIF}: Generative Interpretable Faces},
    author = {Ghosh, Partha and Gupta, Pravir Singh and Uziel, Roy and Ranjan, Anurag and Black, Michael J. and Bolkart, Timo},
    booktitle = {International Conference on 3D Vision (3DV)},
    pages     = {868--878},
    year = {2020},
    url = {http://gif.is.tue.mpg.de/}
}

@InProceedings{bounareli2023hyperreenact,
    author    = {Bounareli, Stella and Tzelepis, Christos and Argyriou, Vasileios and Patras, Ioannis and Tzimiropoulos, Georgios},
    title     = {HyperReenact: One-Shot Reenactment via Jointly Learning to Refine and Retarget Faces},
    booktitle = {Proceedings of the IEEE/CVF International Conference on Computer Vision (ICCV)},
    year      = {2023},
}

@article{ye2023ip,
  title={Ip-adapter: Text compatible image prompt adapter for text-to-image diffusion models},
  author={Ye, Hu and Zhang, Jun and Liu, Sibo and Han, Xiao and Yang, Wei},
  journal={arXiv preprint arXiv:2308.06721},
  year={2023}
}

@article{gal2022image,
  title={An image is worth one word: Personalizing text-to-image generation using textual inversion},
  author={Gal, Rinon and Alaluf, Yuval and Atzmon, Yuval and Patashnik, Or and Bermano, Amit H and Chechik, Gal and Cohen-Or, Daniel},
  journal={arXiv preprint arXiv:2208.01618},
  year={2022}
}

@inproceedings{ruiz2023dreambooth,
  title={Dreambooth: Fine tuning text-to-image diffusion models for subject-driven generation},
  author={Ruiz, Nataniel and Li, Yuanzhen and Jampani, Varun and Pritch, Yael and Rubinstein, Michael and Aberman, Kfir},
  booktitle={Proceedings of the IEEE/CVF conference on computer vision and pattern recognition},
  pages={22500--22510},
  year={2023}
}

@inproceedings{wang2018cosface,
  title={Cosface: Large margin cosine loss for deep face recognition},
  author={Wang, Hao and Wang, Yitong and Zhou, Zheng and Ji, Xing and Gong, Dihong and Zhou, Jingchao and Li, Zhifeng and Liu, Wei},
  booktitle={Proceedings of the IEEE conference on computer vision and pattern recognition},
  pages={5265--5274},
  year={2018}
}

@inproceedings{deng2019arcface,
  title={Arcface: Additive angular margin loss for deep face recognition},
  author={Deng, Jiankang and Guo, Jia and Xue, Niannan and Zafeiriou, Stefanos},
  booktitle={Proceedings of the IEEE/CVF conference on computer vision and pattern recognition},
  pages={4690--4699},
  year={2019}
}

@inproceedings{papantoniou2024arc2face,
  title={Arc2face: A foundation model for id-consistent human faces},
  author={Papantoniou, Foivos Paraperas and Lattas, Alexandros and Moschoglou, Stylianos and Deng, Jiankang and Kainz, Bernhard and Zafeiriou, Stefanos},
  booktitle={European Conference on Computer Vision},
  pages={241--261},
  year={2024},
  organization={Springer}
}

@article{wang2024instantid,
  title={InstantID: Zero-shot Identity-Preserving Generation in Seconds},
  author={Wang, Qixun and Bai, Xu and Wang, Haofan and Qin, Zekui and Chen, Anthony},
  journal={arXiv preprint arXiv:2401.07519},
  year={2024}
}

@inproceedings{ding2023diffusionrig,
  title={Diffusionrig: Learning personalized priors for facial appearance editing},
  author={Ding, Zheng and Zhang, Xuaner and Xia, Zhihao and Jebe, Lars and Tu, Zhuowen and Zhang, Xiuming},
  booktitle={Proceedings of the IEEE/CVF conference on computer vision and pattern recognition},
  pages={12736--12746},
  year={2023}
}

@inproceedings{han2024face,
  title={Face-Adapter for Pre-trained Diffusion Models with Fine-Grained ID and Attribute Control},
  author={Han, Yue and Zhu, Junwei and He, Keke and Chen, Xu and Ge, Yanhao and Li, Wei and Li, Xiangtai and Zhang, Jiangning and Wang, Chengjie and Liu, Yong},
  booktitle={European Conference on Computer Vision},
  pages={20--36},
  year={2024},
  organization={Springer}
}

@inproceedings{drobyshev2024emoportraits,
  title={Emoportraits: Emotion-enhanced multimodal one-shot head avatars},
  author={Drobyshev, Nikita and Casademunt, Antoni Bigata and Vougioukas, Konstantinos and Landgraf, Zoe and Petridis, Stavros and Pantic, Maja},
  booktitle={Proceedings of the IEEE/CVF Conference on Computer Vision and Pattern Recognition},
  pages={8498--8507},
  year={2024}
}

@article{wei2024aniportrait,
  title={Aniportrait: Audio-driven synthesis of photorealistic portrait animation},
  author={Wei, Huawei and Yang, Zejun and Wang, Zhisheng},
  journal={arXiv preprint arXiv:2403.17694},
  year={2024}
}

@article{wang2024v,
  title={V-express: Conditional dropout for progressive training of portrait video generation},
  author={Wang, Cong and Tian, Kuan and Zhang, Jun and Guan, Yonghang and Luo, Feng and Shen, Fei and Jiang, Zhiwei and Gu, Qing and Han, Xiao and Yang, Wei},
  journal={arXiv preprint arXiv:2406.02511},
  year={2024}
}

@inproceedings{ma2024follow,
  title={Follow-your-emoji: Fine-controllable and expressive freestyle portrait animation},
  author={Ma, Yue and Liu, Hongyu and Wang, Hongfa and Pan, Heng and He, Yingqing and Yuan, Junkun and Zeng, Ailing and Cai, Chengfei and Shum, Heung-Yeung and Liu, Wei and others},
  booktitle={SIGGRAPH Asia 2024 Conference Papers},
  pages={1--12},
  year={2024}
}

@inproceedings{boutros2022elasticface,
  title={Elasticface: Elastic margin loss for deep face recognition},
  author={Boutros, Fadi and Damer, Naser and Kirchbuchner, Florian and Kuijper, Arjan},
  booktitle={Proceedings of the IEEE/CVF conference on computer vision and pattern recognition},
  pages={1578--1587},
  year={2022}
}

@article{mai2018reconstruction,
  title={On the reconstruction of face images from deep face templates},
  author={Mai, Guangcan and Cao, Kai and Yuen, Pong C and Jain, Anil K},
  journal={IEEE transactions on pattern analysis and machine intelligence},
  volume={41},
  number={5},
  pages={1188--1202},
  year={2018},
  publisher={IEEE}
}

@inproceedings{razzhigaev2020black,
  title={Black-box face recovery from identity features},
  author={Razzhigaev, Anton and Kireev, Klim and Kaziakhmedov, Edgar and Tursynbek, Nurislam and Petiushko, Aleksandr},
  booktitle={Computer Vision--ECCV 2020 Workshops: Glasgow, UK, August 23--28, 2020, Proceedings, Part V 16},
  pages={462--475},
  year={2020},
  organization={Springer}
}

@inproceedings{vendrow2021realistic,
  title={Realistic face reconstruction from deep embeddings},
  author={Vendrow, Edward and Vendrow, Joshua},
  booktitle={NeurIPS 2021 Workshop Privacy in Machine Learning},
  year={2021}
}

@inproceedings{duong2020vec2face,
  title={Vec2face: Unveil human faces from their blackbox features in face recognition},
  author={Duong, Chi Nhan and Truong, Thanh-Dat and Luu, Khoa and Quach, Kha Gia and Bui, Hung and Roy, Kaushik},
  booktitle={Proceedings of the IEEE/CVF Conference on Computer Vision and Pattern Recognition},
  pages={6132--6141},
  year={2020}
}

@article{truong2022vec2face,
  title={Vec2face-v2: Unveil human faces from their blackbox features via attention-based network in face recognition},
  author={Truong, Thanh-Dat and Duong, Chi Nhan and Le, Ngan and Savvides, Marios and Luu, Khoa},
  journal={arXiv preprint arXiv:2209.04920},
  year={2022}
}

@inproceedings{kansy2023controllable,
  title={Controllable inversion of black-box face recognition models via diffusion},
  author={Kansy, Manuel and Ra{\"e}l, Anton and Mignone, Graziana and Naruniec, Jacek and Schroers, Christopher and Gross, Markus and Weber, Romann M},
  booktitle={Proceedings of the IEEE/CVF International Conference on Computer Vision},
  pages={3167--3177},
  year={2023}
}

@inproceedings{ruiz2024hyperdreambooth,
  title={Hyperdreambooth: Hypernetworks for fast personalization of text-to-image models},
  author={Ruiz, Nataniel and Li, Yuanzhen and Jampani, Varun and Wei, Wei and Hou, Tingbo and Pritch, Yael and Wadhwa, Neal and Rubinstein, Michael and Aberman, Kfir},
  booktitle={Proceedings of the IEEE/CVF conference on computer vision and pattern recognition},
  pages={6527--6536},
  year={2024}
}

@article{hu2022lora,
  title={Lora: Low-rank adaptation of large language models.},
  author={Hu, Edward J and Shen, Yelong and Wallis, Phillip and Allen-Zhu, Zeyuan and Li, Yuanzhi and Wang, Shean and Wang, Lu and Chen, Weizhu and others},
  journal={ICLR},
  volume={1},
  number={2},
  pages={3},
  year={2022}
}

@article{yang2024lora,
  title={Lora-composer: Leveraging low-rank adaptation for multi-concept customization in training-free diffusion models},
  author={Yang, Yang and Wang, Wen and Peng, Liang and Song, Chaotian and Chen, Yao and Li, Hengjia and Yang, Xiaolong and Lu, Qinglin and Cai, Deng and Wu, Boxi and others},
  journal={arXiv preprint arXiv:2403.11627},
  year={2024}
}

@article{yuan2023celebbasis,
          title={Inserting Anybody in Diffusion Models via Celeb Basis},
          author={Yuan, Ge and Cun, Xiaodong and Zhang, Yong and Li, Maomao and Qi, Chenyang and Wang, Xintao and Shan, Ying and Zheng, Huicheng},
          journal={arXiv preprint arXiv:2306.00926},
          year={2023}
        }

@article{wang2024stableidentity,
  title={Stableidentity: Inserting anybody into anywhere at first sight},
  author={Wang, Qinghe and Jia, Xu and Li, Xiaomin and Li, Taiqing and Ma, Liqian and Zhuge, Yunzhi and Lu, Huchuan},
  journal={arXiv preprint arXiv:2401.15975},
  year={2024}
}

@article{pan2023kosmos,
  title={Kosmos-g: Generating images in context with multimodal large language models},
  author={Pan, Xichen and Dong, Li and Huang, Shaohan and Peng, Zhiliang and Chen, Wenhu and Wei, Furu},
  journal={arXiv preprint arXiv:2310.02992},
  year={2023}
}

@article{xiao2024fastcomposer,
  title={Fastcomposer: Tuning-free multi-subject image generation with localized attention},
  author={Xiao, Guangxuan and Yin, Tianwei and Freeman, William T and Durand, Fr{\'e}do and Han, Song},
  journal={International Journal of Computer Vision},
  pages={1--20},
  year={2024},
  publisher={Springer}
}

@article{chen2023photoverse,
  title={Photoverse: Tuning-free image customization with text-to-image diffusion models},
  author={Chen, Li and Zhao, Mengyi and Liu, Yiheng and Ding, Mingxu and Song, Yangyang and Wang, Shizun and Wang, Xu and Yang, Hao and Liu, Jing and Du, Kang and others},
  journal={arXiv preprint arXiv:2309.05793},
  year={2023}
}

@inproceedings{li2024photomaker,
  title={Photomaker: Customizing realistic human photos via stacked id embedding},
  author={Li, Zhen and Cao, Mingdeng and Wang, Xintao and Qi, Zhongang and Cheng, Ming-Ming and Shan, Ying},
  booktitle={Proceedings of the IEEE/CVF conference on computer vision and pattern recognition},
  pages={8640--8650},
  year={2024}
}

@inproceedings{valevski2023face0,
  title={Face0: Instantaneously conditioning a text-to-image model on a face},
  author={Valevski, Dani and Lumen, Danny and Matias, Yossi and Leviathan, Yaniv},
  booktitle={SIGGRAPH Asia 2023 Conference Papers},
  pages={1--10},
  year={2023}
}

@inproceedings{chen2024dreamidentity,
  title={DreamIdentity: enhanced editability for efficient face-identity preserved image generation},
  author={Chen, Zhuowei and Fang, Shancheng and Liu, Wei and He, Qian and Huang, Mengqi and Mao, Zhendong},
  booktitle={Proceedings of the AAAI Conference on Artificial Intelligence},
  volume={38},
  number={2},
  pages={1281--1289},
  year={2024}
}

@inproceedings{peng2024portraitbooth,
  title={Portraitbooth: A versatile portrait model for fast identity-preserved personalization},
  author={Peng, Xu and Zhu, Junwei and Jiang, Boyuan and Tai, Ying and Luo, Donghao and Zhang, Jiangning and Lin, Wei and Jin, Taisong and Wang, Chengjie and Ji, Rongrong},
  booktitle={Proceedings of the IEEE/CVF Conference on Computer Vision and Pattern Recognition},
  pages={27080--27090},
  year={2024}
}

@article{yan2023facestudio,
  title={Facestudio: Put your face everywhere in seconds},
  author={Yan, Yuxuan and Zhang, Chi and Wang, Rui and Zhou, Yichao and Zhang, Gege and Cheng, Pei and Yu, Gang and Fu, Bin},
  journal={arXiv preprint arXiv:2312.02663},
  year={2023}
}

@inproceedings{zhu2021webface260m,
  title={Webface260m: A benchmark unveiling the power of million-scale deep face recognition},
  author={Zhu, Zheng and Huang, Guan and Deng, Jiankang and Ye, Yun and Huang, Junjie and Chen, Xinze and Zhu, Jiagang and Yang, Tian and Lu, Jiwen and Du, Dalong and others},
  booktitle={Proceedings of the IEEE/CVF Conference on Computer Vision and Pattern Recognition},
  pages={10492--10502},
  year={2021}
}

@inproceedings{yin2022styleheat,
  title={Styleheat: One-shot high-resolution editable talking face generation via pre-trained stylegan},
  author={Yin, Fei and Zhang, Yong and Cun, Xiaodong and Cao, Mingdeng and Fan, Yanbo and Wang, Xuan and Bai, Qingyan and Wu, Baoyuan and Wang, Jue and Yang, Yujiu},
  booktitle={European conference on computer vision},
  pages={85--101},
  year={2022},
  organization={Springer}
}

@article{kim2018deep,
  title={Deep video portraits},
  author={Kim, Hyeongwoo and Garrido, Pablo and Tewari, Ayush and Xu, Weipeng and Thies, Justus and Niessner, Matthias and P{\'e}rez, Patrick and Richardt, Christian and Zollh{\"o}fer, Michael and Theobalt, Christian},
  journal={ACM transactions on graphics (TOG)},
  volume={37},
  number={4},
  pages={1--14},
  year={2018},
  publisher={ACM New York, NY, USA}
}

@inproceedings{kirschstein2024diffusionavatars,
  title={Diffusionavatars: Deferred diffusion for high-fidelity 3d head avatars},
  author={Kirschstein, Tobias and Giebenhain, Simon and Nie{\ss}ner, Matthias},
  booktitle={Proceedings of the IEEE/CVF Conference on Computer Vision and Pattern Recognition},
  pages={5481--5492},
  year={2024}
}

@inproceedings{khakhulin2022realistic,
  title={Realistic one-shot mesh-based head avatars},
  author={Khakhulin, Taras and Sklyarova, Vanessa and Lempitsky, Victor and Zakharov, Egor},
  booktitle={European Conference on Computer Vision},
  pages={345--362},
  year={2022},
  organization={Springer}
}

@inproceedings{hu2024animate,
  title={Animate anyone: Consistent and controllable image-to-video synthesis for character animation},
  author={Hu, Li},
  booktitle={Proceedings of the IEEE/CVF Conference on Computer Vision and Pattern Recognition},
  pages={8153--8163},
  year={2024}
}

@article{yang2023diffusion,
  title={Diffusion models: A comprehensive survey of methods and applications},
  author={Yang, Ling and Zhang, Zhilong and Song, Yang and Hong, Shenda and Xu, Runsheng and Zhao, Yue and Zhang, Wentao and Cui, Bin and Yang, Ming-Hsuan},
  journal={ACM Computing Surveys},
  volume={56},
  number={4},
  pages={1--39},
  year={2023},
  publisher={ACM New York, NY, USA}
}

@inproceedings{rombach2022high,
  title={High-resolution image synthesis with latent diffusion models},
  author={Rombach, Robin and Blattmann, Andreas and Lorenz, Dominik and Esser, Patrick and Ommer, Bj{\"o}rn},
  booktitle={Proceedings of the IEEE/CVF conference on computer vision and pattern recognition},
  pages={10684--10695},
  year={2022}
}

@article{ho2020denoising,
  title={Denoising diffusion probabilistic models},
  author={Ho, Jonathan and Jain, Ajay and Abbeel, Pieter},
  journal={Advances in neural information processing systems},
  volume={33},
  pages={6840--6851},
  year={2020}
}

@inproceedings{bar2024lumiere,
  title={Lumiere: A space-time diffusion model for video generation},
  author={Bar-Tal, Omer and Chefer, Hila and Tov, Omer and Herrmann, Charles and Paiss, Roni and Zada, Shiran and Ephrat, Ariel and Hur, Junhwa and Liu, Guanghui and Raj, Amit and others},
  booktitle={SIGGRAPH Asia 2024 Conference Papers},
  pages={1--11},
  year={2024}
}

@article{blattmann2023stable,
  title={Stable video diffusion: Scaling latent video diffusion models to large datasets},
  author={Blattmann, Andreas and Dockhorn, Tim and Kulal, Sumith and Mendelevitch, Daniel and Kilian, Maciej and Lorenz, Dominik and Levi, Yam and English, Zion and Voleti, Vikram and Letts, Adam and others},
  journal={arXiv preprint arXiv:2311.15127},
  year={2023}
}

@article{li2017learning,
  title={Learning a model of facial shape and expression from 4D scans.},
  author={Li, Tianye and Bolkart, Timo and Black, Michael J and Li, Hao and Romero, Javier},
  journal={ACM Trans. Graph.},
  volume={36},
  number={6},
  pages={194--1},
  year={2017}
}

@inproceedings{giebenhain2023learning,
  title={Learning neural parametric head models},
  author={Giebenhain, Simon and Kirschstein, Tobias and Georgopoulos, Markos and R{\"u}nz, Martin and Agapito, Lourdes and Nie{\ss}ner, Matthias},
  booktitle={Proceedings of the IEEE/CVF Conference on Computer Vision and Pattern Recognition},
  pages={21003--21012},
  year={2023}
}

@inproceedings{chung2024style,
  title={Style injection in diffusion: A training-free approach for adapting large-scale diffusion models for style transfer},
  author={Chung, Jiwoo and Hyun, Sangeek and Heo, Jae-Pil},
  booktitle={Proceedings of the IEEE/CVF conference on computer vision and pattern recognition},
  pages={8795--8805},
  year={2024}
}

@inproceedings{huang2017arbitrary,
  title={Arbitrary style transfer in real-time with adaptive instance normalization},
  author={Huang, Xun and Belongie, Serge},
  booktitle={Proceedings of the IEEE international conference on computer vision},
  pages={1501--1510},
  year={2017}
}

@inproceedings{deng2019accurate,
  title={Accurate 3d face reconstruction with weakly-supervised learning: From single image to image set},
  author={Deng, Yu and Yang, Jiaolong and Xu, Sicheng and Chen, Dong and Jia, Yunde and Tong, Xin},
  booktitle={Proceedings of the IEEE/CVF conference on computer vision and pattern recognition workshops},
  pages={0--0},
  year={2019}
}

@inproceedings{wright2022artfid,
  title={Artfid: Quantitative evaluation of neural style transfer},
  author={Wright, Matthias and Ommer, Bj{\"o}rn},
  booktitle={DAGM German Conference on Pattern Recognition},
  pages={560--576},
  year={2022},
  organization={Springer}
}

@inproceedings{karras2019style,
  title={A style-based generator architecture for generative adversarial networks},
  author={Karras, Tero and Laine, Samuli and Aila, Timo},
  booktitle={Proceedings of the IEEE/CVF conference on computer vision and pattern recognition},
  pages={4401--4410},
  year={2019}
}

@article{nagrani2020voxceleb,
  title={Voxceleb: Large-scale speaker verification in the wild},
  author={Nagrani, Arsha and Chung, Joon Son and Xie, Weidi and Zisserman, Andrew},
  journal={Computer Speech \& Language},
  volume={60},
  pages={101027},
  year={2020},
  publisher={Elsevier}
}

@article{chung2018voxceleb2,
  title={Voxceleb2: Deep speaker recognition},
  author={Chung, Joon Son and Nagrani, Arsha and Zisserman, Andrew},
  journal={arXiv preprint arXiv:1806.05622},
  year={2018}
}

@inproceedings{zheng2022general,
  title={General facial representation learning in a visual-linguistic manner},
  author={Zheng, Yinglin and Yang, Hao and Zhang, Ting and Bao, Jianmin and Chen, Dongdong and Huang, Yangyu and Yuan, Lu and Chen, Dong and Zeng, Ming and Wen, Fang},
  booktitle={Proceedings of the IEEE/CVF conference on computer vision and pattern recognition},
  pages={18697--18709},
  year={2022}
}

@article{ye2023geneface,
  title={Geneface: Generalized and high-fidelity audio-driven 3d talking face synthesis},
  author={Ye, Zhenhui and Jiang, Ziyue and Ren, Yi and Liu, Jinglin and He, Jinzheng and Zhao, Zhou},
  journal={arXiv preprint arXiv:2301.13430},
  year={2023}
}

@inproceedings{chandran2021adaptive,
  title={Adaptive convolutions for structure-aware style transfer},
  author={Chandran, Prashanth and Zoss, Gaspard and Gotardo, Paulo and Gross, Markus and Bradley, Derek},
  booktitle={Proceedings of the IEEE/CVF conference on computer vision and pattern recognition},
  pages={7972--7981},
  year={2021}
}

@inproceedings{liu2021adaattn,
  title={Adaattn: Revisit attention mechanism in arbitrary neural style transfer},
  author={Liu, Songhua and Lin, Tianwei and He, Dongliang and Li, Fu and Wang, Meiling and Li, Xin and Sun, Zhengxing and Li, Qian and Ding, Errui},
  booktitle={Proceedings of the IEEE/CVF international conference on computer vision},
  pages={6649--6658},
  year={2021}
}

@article{jeong2023training,
  title={Training-free style transfer emerges from h-space in diffusion models},
  author={Jeong, Jaeseok and Kwon, Mingi and Uh, Youngjung},
  journal={arXiv preprint arXiv:2303.15403},
  volume={3},
  number={1},
  pages={2},
  year={2023}
}
